%% file: main.tex
\documentclass[12pt,oneside,openany]{report}
\usepackage[a4paper,width=150mm,top=25mm,bottom=25mm]{geometry}
\usepackage[english]{babel}
\usepackage[T1]{fontenc}
\usepackage{lmodern}
\usepackage{csquotes} 
\usepackage{amsmath} 
\usepackage{subcaption}
\usepackage{graphicx} 
\graphicspath{{images/}} 
\usepackage{caption} 
\usepackage{array} 
\usepackage{microtype}
\usepackage[nottoc]{tocbibind} 
\usepackage{multirow}
\usepackage[normalem]{ulem} 
\usepackage{float}
\usepackage{hyperref} 
\hypersetup{
    colorlinks=true,
    linkcolor=black,
    filecolor=magenta,      
    urlcolor=black,
    citecolor=black,
}
\usepackage{nomencl} 
\renewcommand{\nompreamble}{The next list describes several symbols \& abbreviation that will be later used within the body of the document}
\makenomenclature

\usepackage[backend=biber,style=ieee,sorting=ynt]{biblatex} 
\let\cleardoublepage=\clearpage 

\begin{document}
\thispagestyle{empty} 
\input{core/titlepage} 
\cleardoublepage

\pagenumbering{roman} 

\phantomsection
\addcontentsline{toc}{chapter}{Declaration}
\input{core/declaration}

\phantomsection
\addcontentsline{toc}{chapter}{Approval}
\input{core/approval}


\phantomsection
\addcontentsline{toc}{chapter}{Abstract}
\input{core/abstract}

\phantomsection
\addcontentsline{toc}{chapter}{Acknowledgement}
\input{core/acknowledgement}

\renewcommand{\contentsname}{Table of Contents} 
\cleardoublepage
\phantomsection
\addcontentsline{toc}{chapter}{Table of Contents} 
\tableofcontents 

\listoffigures 
\listoftables 

\printnomenclature 
\addcontentsline{toc}{chapter}{Nomenclature}
\cleardoublepage

\pagenumbering{arabic} 

\chapter{Introduction}
\input{chapters/chapter_1.tex}

\chapter{Literature Review}
\input{chapters/chapter_2.tex}

\chapter{Methodology}
\input{chapters/chapter_3.tex}
\chapter{Results and Discussion}
\input{chapters/chapter_4.tex}

\chapter{Conclusion}
\input{chapters/chapter_5.tex}

\phantomsection

\printbibliography 
\addcontentsline{toc}{chapter}{Bibliography}


\clearpage
\phantomsection
\addcontentsline{toc}{chapter}{LLM-as-a-Judge Prompt}
\section*{LLM-as-a-Judge Prompt}
\label{sec:prompt}
\input{chapters/chapter_7.tex}
\end{document}

%% file: core/titlepage.tex
\input{core/commands}

\begin{titlepage}
\renewcommand*{\thepage}{Title} 
    \begin{center}
        \vspace*{3cm} 
        \vspace{1cm}
        {\fontsize{16pt}{22pt}\selectfont{Automated Species Identification in Camera Trap Images for Wildlife Conservation}
        } 
        
        \vspace{1.5cm}
        
        \text{by}
        
        \vspace{0.5cm}
            \color{\studentonecolor}
        	\studentonename\\\studentoneid\\
        	\color{\studenttwocolor}
	        \studenttwoname\\\studenttwoid\\
	        \color{\studentthreecolor}
	        \studentthreename\\\studentthreeid\\
	        \color{\studentfourcolor}
	        \studentfourname\\\studentfourid\\
	        \color{\studentfivecolor}
	        \studentfivename\\\studentfiveid\\ 
            \color{black}
        \vspace{1.5cm}
        
        	A thesis submitted to the Department of Computer Science and Engineering\\
            in partial fulfillment of the requirements for the degree of\\
            B.Sc. in Computer Science

        \vspace{2.5cm}
        
    		Department of Computer Science and Engineering\\
            Brac University\\
            June 2025
        
        \vspace{3cm}
        
    		\copyright\ 2025. Brac University\\
            All rights reserved.
    
    \end{center}

\end{titlepage}

%% file: core/commands.tex
\newcommand*{\studentoneid}{21201234}
\newcommand*{\studentonename}{Nowshin Amin}
\newcommand*{\studenttwoid}{21201179}
\newcommand*{\studenttwoname}{Nafisa Tabassum Oyshi}
\newcommand*{\studentthreeid}{21201056}
\newcommand*{\studentthreename}{Tahmid Abrar Zidan}
\newcommand*{\studentfourid}{21241021}
\newcommand*{\studentfourname}{Miftaun Noor}
\newcommand*{\studentfiveid}{21201080}
\newcommand*{\studentfivename}{Md. Abrar Rahman Shafin}
\newcommand*{\studentonecolor}{black}
\newcommand*{\studenttwocolor}{black}
\newcommand*{\studentthreecolor}{black}
\newcommand*{\studentfourcolor}{black}
\newcommand*{\studentfivecolor}{black}

%% file: core/declaration.tex
\newcommand*\wildcard[2][6cm]{\vspace{1cm}\parbox{#1}{\hrulefill\par#2}} 


\section*{Declaration}

It is hereby declared that

\begin{enumerate} 
  \item The thesis submitted is our own original work while completing degree at Brac University.
  \item The thesis does not contain material previously published or written by a third party, except where this is appropriately cited through full and accurate referencing.
  \item The thesis does not contain material which has been accepted, or submitted, for any other degree or diploma at a university or other institution.
  \item We have acknowledged all main sources of help.
\end{enumerate}

\vspace{1cm}
\textbf{Student’s Full Name \& Signature:} 
\vspace{.5cm}

\begin{center}
    \begin{tabular}[b]{@{} p{6cm} @{}}
        \includegraphics[width=6cm]{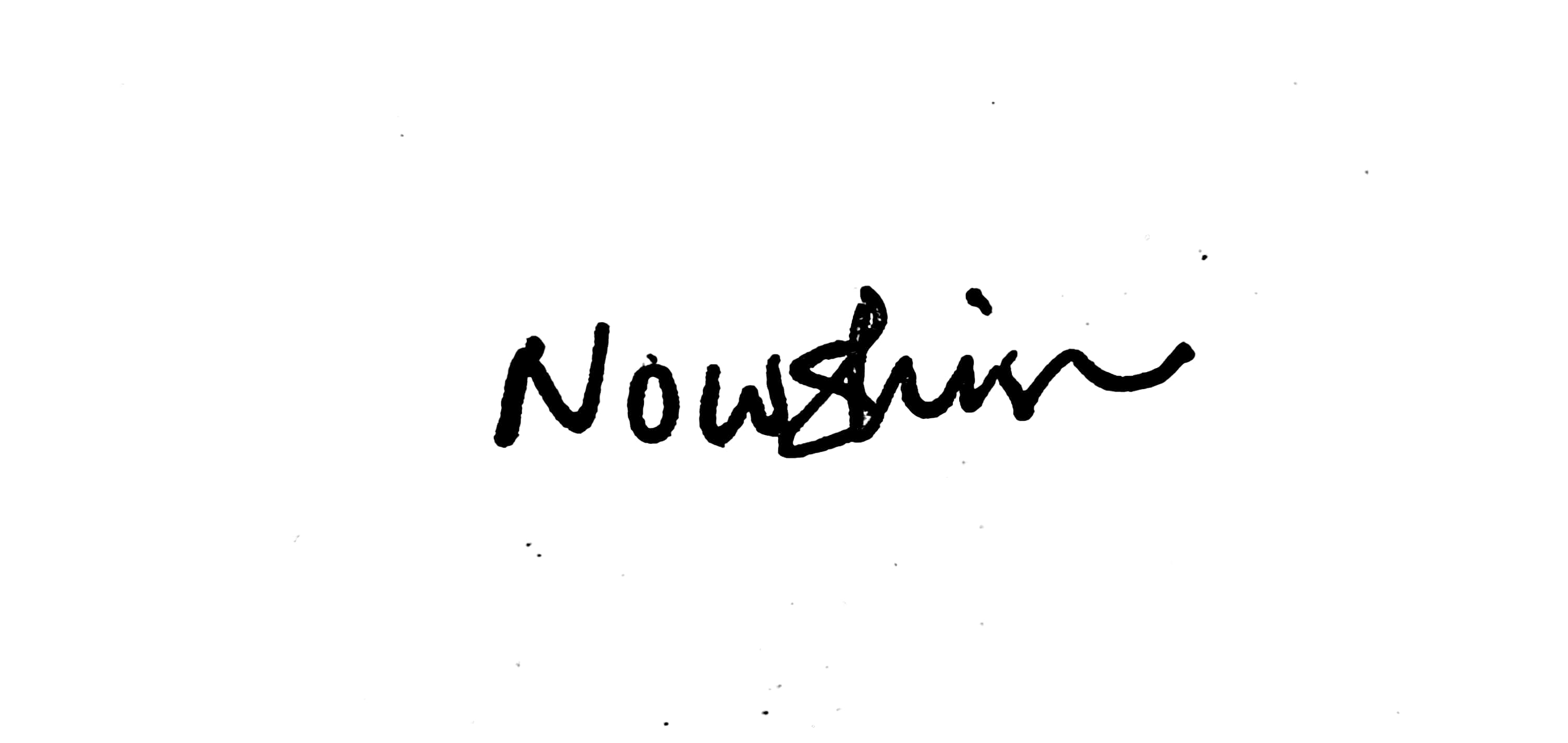} \\
        \hline
        \centerline{\studentonename}
        \centerline{\studentoneid}
    \end{tabular}
    \hspace{2.5cm} 
    \begin{tabular}[b]{@{} p{6cm} @{}}
        \includegraphics[width=6cm]{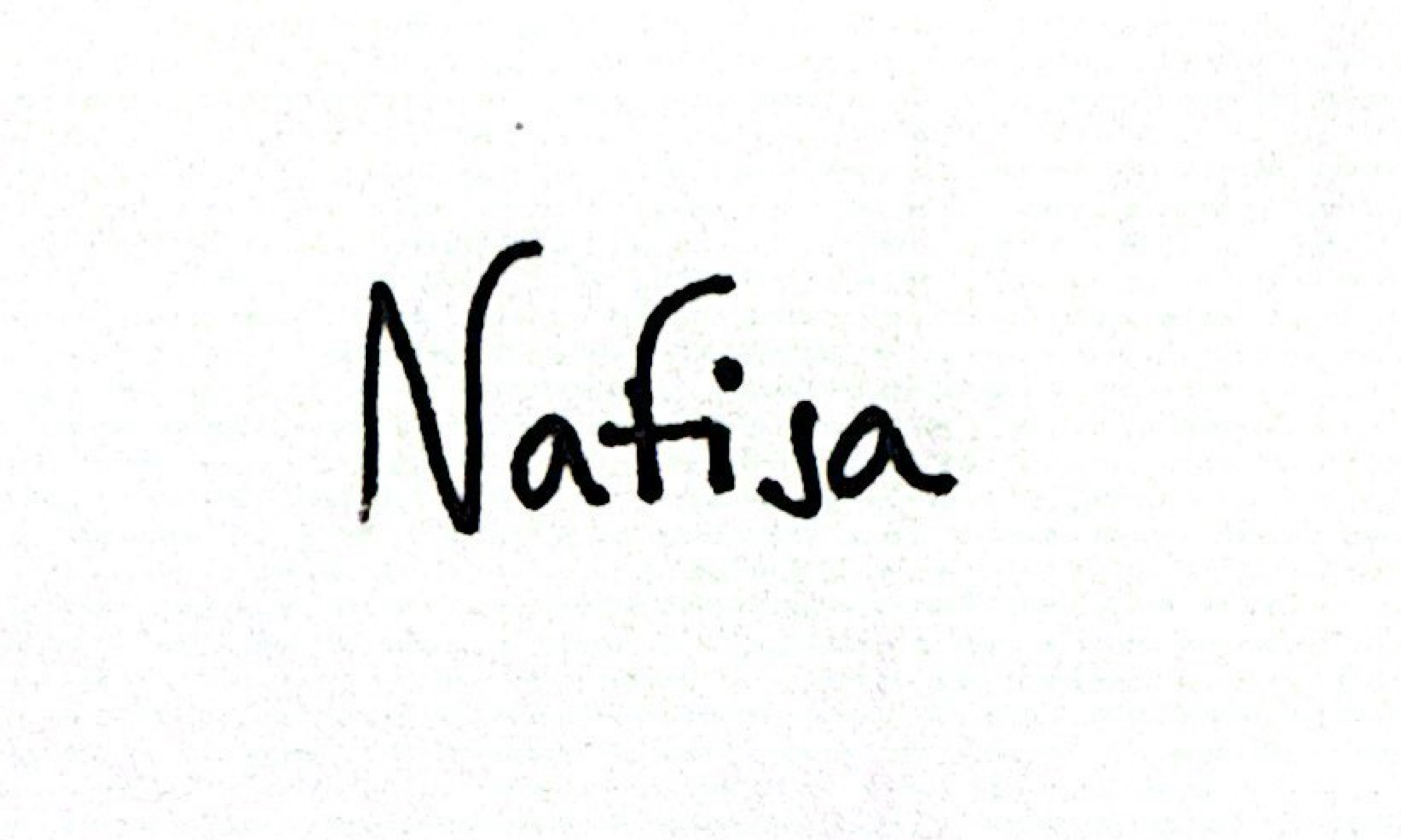} \\
        \hline
        \centerline{\studenttwoname}
        \centerline{\studenttwoid}
    \end{tabular} \\ 
    \begin{tabular}[b]{@{} p{6cm} @{}}
        \includegraphics[width=6cm]{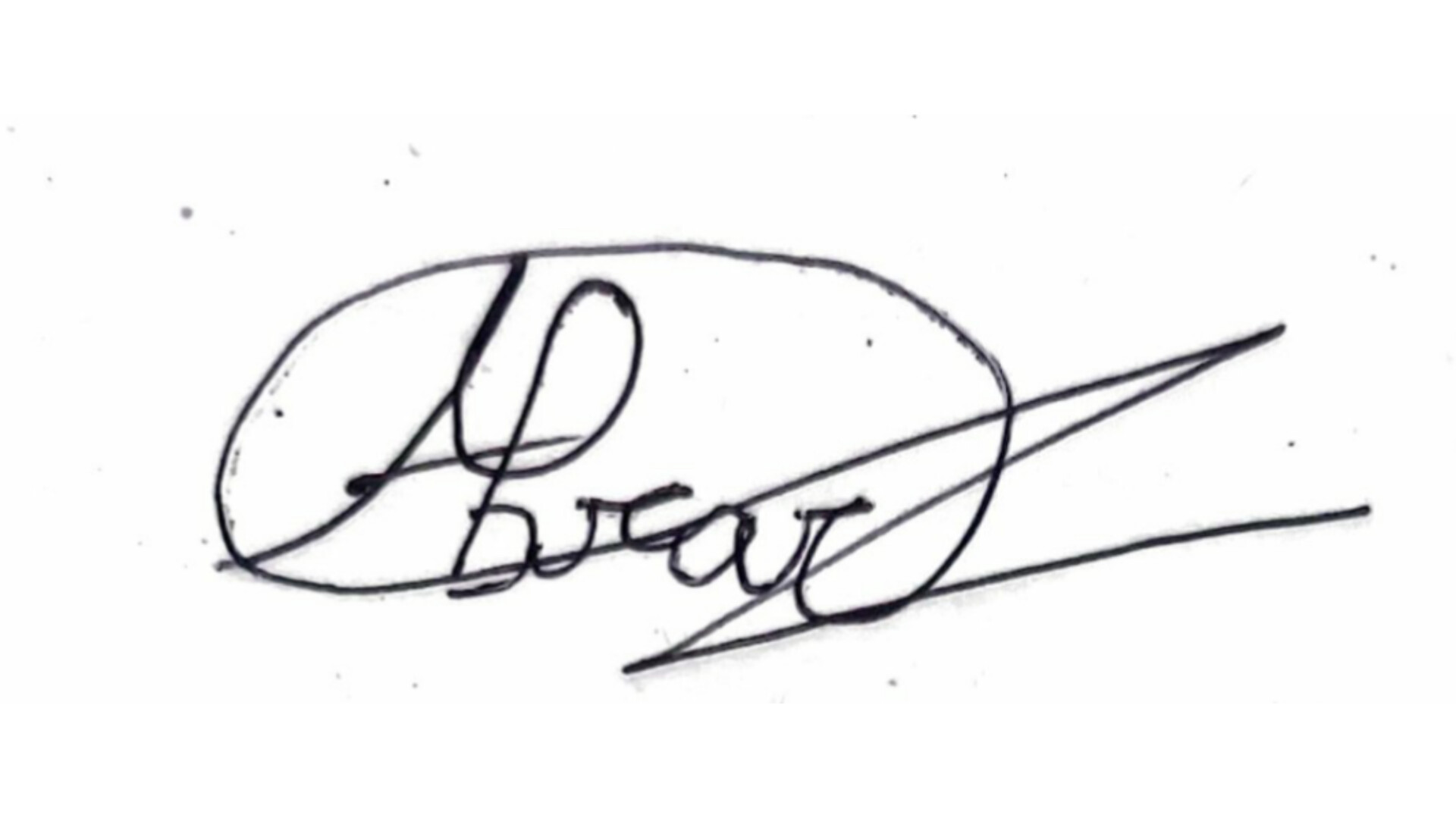} \\
        \hline
        \centerline{\studentthreename}
        \centerline{\studentthreeid}
    \end{tabular}
    \hspace{2.5cm}
    \begin{tabular}[b]{@{} p{6cm} @{}}
        \includegraphics[width=6cm]{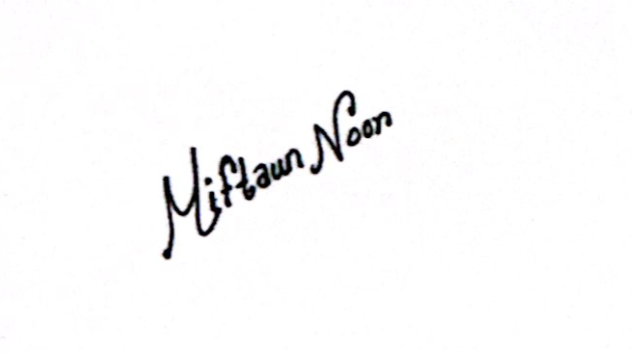} \\
        \hline
        \centerline{\studentfourname}
        \centerline{\studentfourid}
    \end{tabular} \\
    \begin{tabular}[b]{@{} p{6cm} @{}}
        \includegraphics[width=6cm]{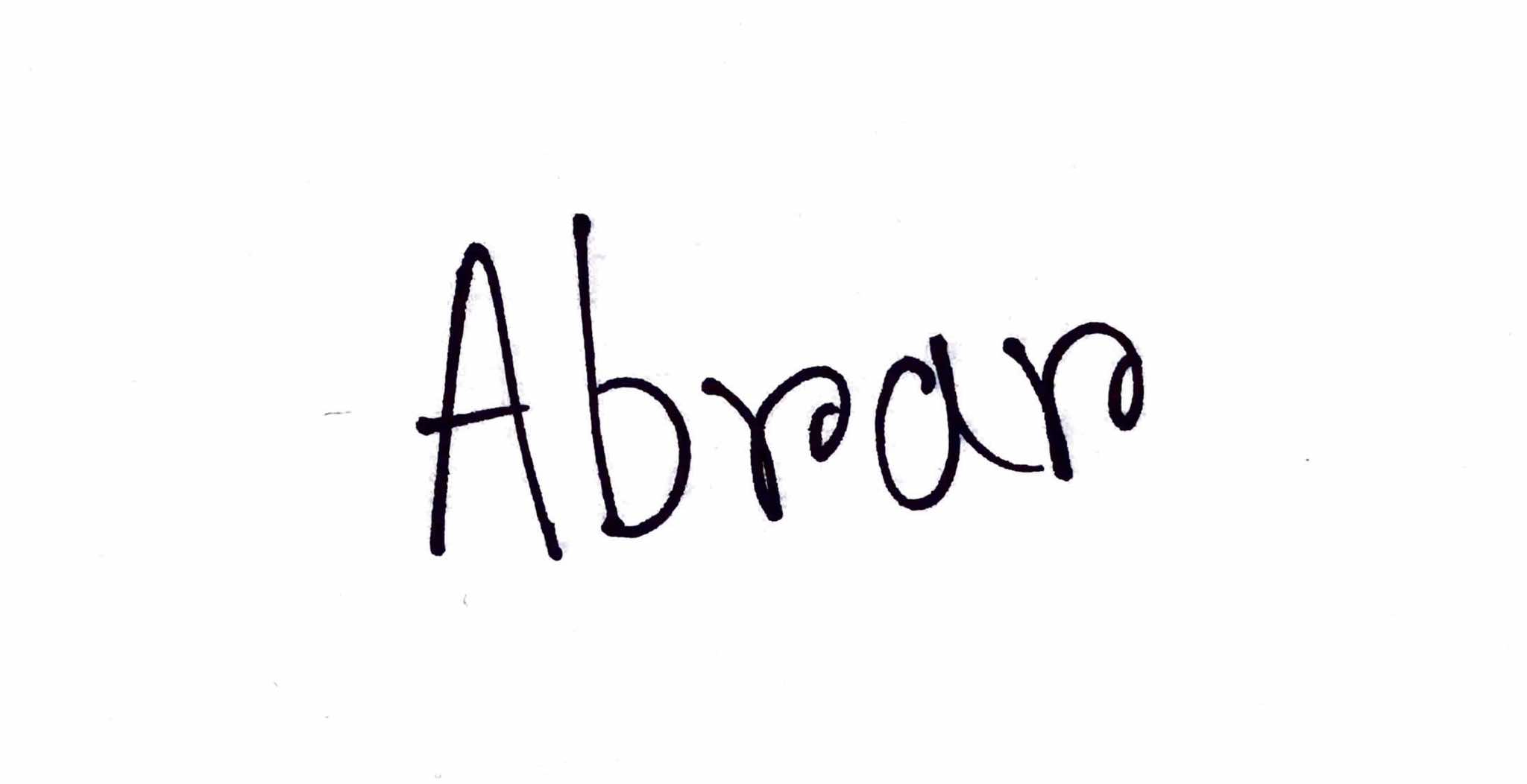} \\
        \hline
        \centerline{\studentfivename}
        \centerline{\studentfiveid}
    \end{tabular}
\end{center}

\pagebreak

%% file: core/approval.tex
\section*{Approval}



\vspace{0.5cm}

\vspace{1cm}

Supervisor:\\
(Member)
\begin{center}
    \hspace{6cm}\includegraphics[width=3cm]{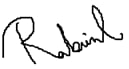} \\
    \hspace{7cm} \wildcard{\centerline{Md. Golam Rabiul Alam, PhD}
    \centerline{Professor}
    \centerline{Department of Computer Science and Engineering}
    \centerline{Brac University} } \hspace{1cm} 
\end{center}

Co-Supervisor:\\
(Member)
\begin{center}
    \hspace{6cm}\includegraphics[width=4cm]{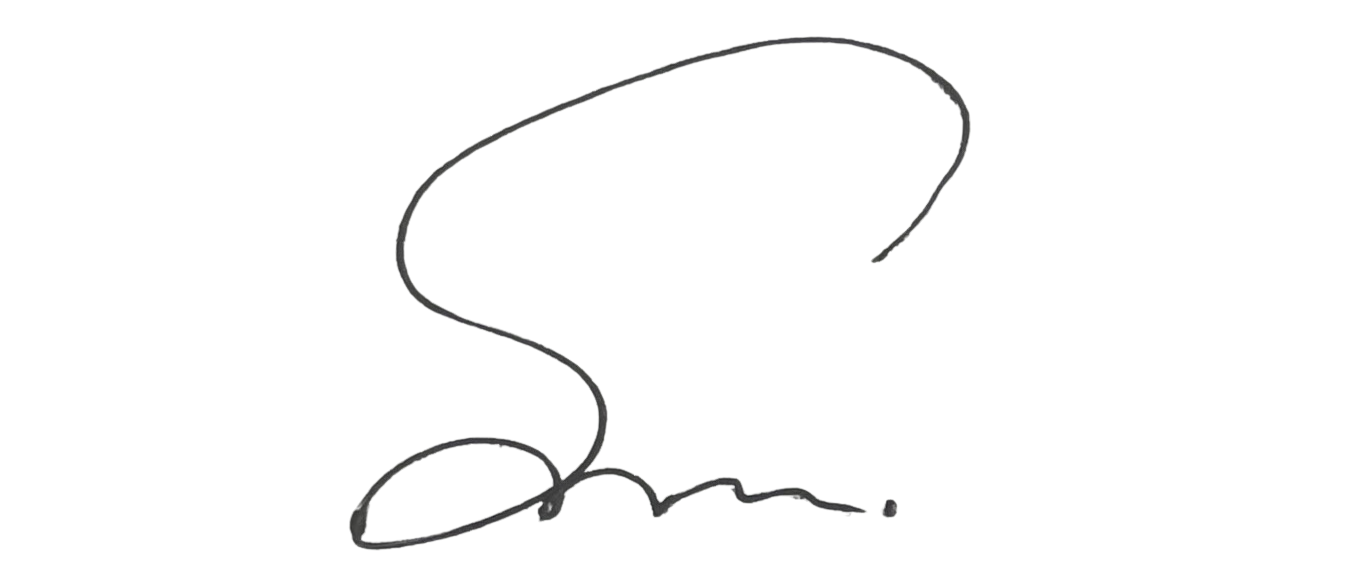} \\
    \hspace{7cm} \wildcard{\centerline{Md Sabbir Ahmed} 
    \centerline{Lecturer}
    \centerline{Department of Computer Science and Engineering}
    \centerline{Brac University} } \hspace{1cm} 

\end{center}

Head of Department:\\
(Chair)
\begin{center}
    \hspace{7cm} \wildcard{\centerline{Sadia Hamid Kazi, PhD}
    \centerline{Chairperson and Associate Professor}
    \centerline{Department of Computer Science and Engineering }
    \centerline{Brac University} } \hspace{1cm} 
\end{center}

\pagebreak

%% file: core/abstract.tex
\section*{Abstract}
Wildlife conservation involves protecting, preserving, and managing wildlife species and their habitats. With today’s rapid pace of human development, climate change, and other unsustainable practices, the need for wildlife conservation has heightened. Despite significant progress in species identification using deep-learning models, significant challenges still remain in effectively detecting small animals in low-contrast trap images due to limited feature extraction capabilities. This thesis presents a novel end-to-end framework integrating a self-attention mechanism to address these limitations. The proposed architecture involves a Swin-BiFPN backbone integrated in a Faster RCNN detection network, coupled with a visual semantic extraction module driven by the LLaVA v1.5 (13B) multimodal large language model. The detection framework, capable of extracting crucial features in challenging trap images, demonstrates consistently high results and robust generalization capabilities. Furthermore, the visual semantic extraction module provides zero-shot detection capability, as well as providing valuable insights and emergent cues of the animal’s behavior, further supporting the conservation effort. The  MLLM evaluation was conducted using both traditional NLP metrics (precision, recall, F1, and SBERT similarity) and subjective scoring by LLM-based judges (GPT-4.1 and GROK 3.0), across five MLLMs, demonstrating the model’s strong performance in visual description generation. The proposed framework improves detection accuracy across low-contrast trap images and small animals while also demonstrating zero-shot detection capability leveraging the MLLM.

\vspace{1cm}
\textbf{Keywords:} Wildlife Conservation, Deep learning, CNN, Swin Transformer, Bi-FPN, Faster-RCNN, MLLM, Animal Identification, Camera Trap Images
\pagebreak

%% file: core/acknowledgement.tex
\section*{Acknowledgement}
We would like to express our sincere gratitude towards everyone who have helped us to conclude our thesis work.\\[0.1cm]

Firstly, we would like to thank the Almighty Allah for his countless blessings and strength that enabled us to complete this research. Without His grace and mercy this achievement would not have been possible.\\[0.1cm]

We are deeply grateful to our supervisor Dr. Md. Golam Rabiul Alam, for his invaluable guidance and patience throughout this research journey. We are also profoundly thankful to our co-supervisor Md. Sabbir Ahmed, for his continuous support and mentorship for the entire duration of this research. Their expertise, constructive feedback and encouragement have been instrumental in shaping this study.\\[0.1cm]

We would also like to extend our gratitude towards our lab technical officer, Md. Injamam Hossain for his prompt responsiveness throughout the research work.\\[0.1cm]

Lastly, we are thankful to the Department of Computer Science and Engineering, BRAC University for providing us with the opportunity and resources to conduct this study.

%% file: chapters/chapter_1.tex
Before humans had a significant impact on the environment, animals could travel freely through woods, mountains, and rivers. However, as time passed, the natural order started to change. For food, people went animal hunting, chopped down trees for farming, and created cities in the places where forests once were. Many species began to go extinct as these changes persisted. The realization that animals and their habitats needed to be protected before it was too late began with this. This is the origin of the concept of Animal conservation, a movement aimed at preventing the extinction of species.\\[0.1cm]

In Bangladesh, the call for conservation came a little later. Elephants roaming the highlands and the powerful Bengal tiger stalking the Sundarbans are just two examples of the amazing biodiversity that this country is endowed with. However, the habitats of many animals started to become smaller as the population increased and more land was utilized for urbanization and agriculture. Rivers became contaminated, forests were cleared, and creatures that had previously flourished now faced new threats. As more people became aware of this loss, the conservation movement gradually spread to Bangladesh.\\[0.1cm]

With its thick mangrove forests, the Sundarbans rose to prominence as a conservation icon. The Bengal tiger, one of the most famous and endangered creatures in the world, lives in this jungle, which is also a UNESCO World Heritage Site \cite{unescoSundarbans}. The tigers became the focal point of Bangladesh's conservation efforts, along with other animals including freshwater dolphins and elephants. These initiatives aimed to preserve the natural equilibrium as much as to protect animals for the sake of it. Entire ecosystems would collapse in the absence of these species, with disastrous effects on both the environment and humans.\\[0.1cm]

In Bangladesh, the Wildlife (Conservation \& Security) Act of 2012 is a key legal document for wildlife protection, building on earlier efforts to regulate hunting and establish protected areas for species like the Bengal tiger, elephants, and others \cite{bangladesh2017forestry}. Before this, conservation efforts were more scattered, but the act created a more structured and regulated approach to wildlife preservation.\\[0.1cm]

However, conservation is a difficult task. It's quite difficult to monitor animals in the field, comprehend their behavior, and determine how many of them are left. This is where technology started to become quite important. Camera traps were invented; these were tiny, covert cameras positioned throughout the wild to take pictures of passing animals. Without upsetting the animals, these photos allowed biologists and researchers to see a look into their life.\\[0.1cm]

Nevertheless, even with the introduction of camera traps, a new problem surfaced. Even with thousands of pictures, it was difficult and time-consuming to identify every species by looking over each image one by one. Is there a way to make this process go more quickly? What if technology could identify animals from these pictures on its own? This idea led to the development of automated species identification systems that make use of artificial intelligence and machine learning. These technologies would quickly identify which species had been captured, saving conservationists time and enabling more accurate research.\\[0.1cm]
\section{Research Problem }
In Bangladesh, deforestation and poaching happen to have long been quite significant threats to wildlife. Between 2001 and 2018, Bangladesh lost (approximately) 6,388 hectares (15,785 acres) of forest in moderately protected areas such as Chunati Wildlife Sanctuary, Baroiyadhala National Park, Hazarikhil Wildlife Sanctuary, and the Dudpukuria-Dhopachari Wildlife Sanctuary. In the Chunati sanctuary alone, deforestation has been increased by a factor of 21 by 2018, with a loss of 502 hectares (1,240 acres) within its boundaries \cite{mongabayMuchBangladeshs}. Further, poaching is a critical issue for wildlife conservation in Bangladesh, threatening species such as tigers, elephants, and various birds. Although exact figures for all species are challenging to ascertain, it's reported that reptiles, including freshwater turtles and tortoises, account for nearly 47.8\% of the wildlife trade incidents domestically \cite{WCS2018}. The illegal trade of species like the Chinese pangolin, jungle cats, and various turtle species is alarmingly high, contributing to their potential local extinction without intervention \cite{Islam2023}.\\[0.1cm]

All of these serve as simple indications to the fact that the condition of the wildlife in our country is in a position of disadvantage and that it clearly demands some actions. In fact, there have been numerous instances of animals moving away from their habitats due to deforestation,  poaching,  and habitat fragmentation, which force them to seek food and safety in unfamiliar areas. Additionally, environmental changes such as climate change as well as, food scarcity further exacerbate their displacement, leading to increased human and wildlife conflicts as animals venture closer to human settlements in search of resources and, it often results in the killing of these animals due to fear, retaliation, or perceived threats to the livestock and the crops.\\[0.1cm]

A few steps have been taken to address wildlife conservation challenges, including the use of technology for animal identification, such as voice recognition systems to monitor species and then track their movements. Additionally, initiatives like community-based conservation programs and also the establishment of protected areas aim to mitigate human-wildlife conflicts and preserve biodiversity. Even though these steps do bring about some progress, they may not be sufficient in the long run which acted as an apparent drive for us to explore advanced methods for ensuring that animals could at least be brought back to their natural habitat once they move out from there.\\[0.1cm]

It has been observed that although steps like voice recognition and protected areas show some progress in wildlife conservation, they fall short when compared to automated animal detection methods used worldwide. Detection through trap images has been successfully implemented in various regions, however, in Bangladesh, trap cameras are very rarely set up and are primarily the result of personal initiatives, making them largely insufficient for comprehensive wildlife monitoring. Having said that, the use of trap images happens to be on the rise and a system to ensure that wild animals can be detected through the aforesaid trap images taken is something that could be capitalized on.
\\[0.1cm]

But then again, developing a system for detecting animals from trap images in Bangladesh poses significant challenges, particularly in collecting and implementing vast datasets. Factors such as limited access to high-quality images, variations in animal appearances due to different environmental conditions, and the necessity for extensive manual labeling make data collection extremely arduous.  Additionally, the lack of established protocols and infrastructure for systematic data gathering seems to further complicate the implementation of an effective detection system.\\[0.1cm]

Keeping those aside, several models could be considered for animal detection in Bangladesh, but each comes with its own set of limitations in this context. EfficientNet, for example, is known for its scalability and accuracy in detecting objects, but then again it struggles with smaller datasets and requires quite a significant computational power \cite{tan2020efficientdet}. Faster R-CNN offers high accuracy but then is slower compared to other models and requires substantial labeled training data, which is often scarce in regions like Bangladesh \cite{10.1007/978-3-030-92632-8_36}. YOLOv11, while faster and capable of real time detection, has been known to show increased issues with overfitting, especially in environments with dense foliage or even cluttered backgrounds. Despite newer versions of detection models emerging, improvements in accuracy seem to remain incremental. Overfitting and false positives, especially in highly variable environments, have increased, limiting their effectiveness in the real world settings unfortunately.\\[0.1cm]

Given these challenges, it is hypothesized that implementing an improved model by the help of incorporating advanced techniques could significantly improve its effectiveness in detecting animals especially, in low-contrast trap images. By combining advanced neural network architectures with supporting layers, this study proposes that the model can address some of these limitations, especially in complex environments where small animals are difficult to distinguish from the background. These additions can enhance the model’s capability to focus on relevant features in the image while ignoring the background clutter, thus helping in reducing false positives. Furthermore, incorporating techniques like deformable convolutions and advanced loss functions such as CIoU Loss can significantly help in improving the accuracy of bounding box predictions and also reducing overfitting issues. Ultimately, this combined approach offers the potential to create a more effective model for animal detection in Bangladesh, capable of addressing the specific challenges posed by this local environment in the spotlight. However,  this also depends on the availability of proper datasets and the resources required for implementing such modifications.\\[0.1cm]

At the same time, the identification of an animal is not enough. Rather, further steps need to be taken so that it can be utilized through a system that would ensure a robust framework capable of identifying animal behavior and act as a fallback option when the detection framework fails.\\[0.1cm]

With all of that into consideration, the specific question this research seeks to answer is: \textbf{How can a fusion model be introduced that effectively detects wildlife in complex environments with high precision while accelerating the process and making way for developing a system?
}\\[0.1cm]

By investigating certain modifications such as improving small animal detection under low-light conditions and refining the bounding box accuracy for various animal postures, while optimizing the computational cost, the following research will contribute to a more robust framework for animal detection. The ultimate goal is to provide a practical solution for wildlife conservation efforts, enabling more effective monitoring as well as rescuing of animals that stray from their very own natural habitats.\\[0.2cm]

\section{Research Objectives}
In order to promote animal conservation initiatives, this thesis aims to create a strong deep-learning fusion model to identify species from camera trap photos. Because there is a shortage of data on small species, which results in low detection accuracy for small species, most models that classify species in camera trap photos are typically not robust enough to handle these class imbalances. Furthermore, without proper pre-processed images from complex environments like shadows or dense vegetation and fine-tuned models, results are often false positives for small species. As further research suggests, small animals are often not detected due to a poor feature extraction method, where the model often overlooks the regions where small animals are present. Additionally, relatively few studies have focused on deriving meaningful insights into animal behavior, as most existing work primarily concentrates on detection tasks. Focusing on these limitations, the objectives of our thesis are: 
\begin{enumerate}
\item  To improve wildlife conservation efforts by developing a model that elevates species identification and classification, enabling more efficient and accurate analysis of rare and endangered animals from camera trap data.     
\item  To address class imbalance for small species by employing oversampling while reducing overfitting issues, data augmentation, and synthetic image generation techniques, enhancing the model’s ability to detect small species.
\item  To achieve a robust model with better generalization ability across diverse wildlife trap images by integrating an attention-based mechanism.
\item  To improve the model's efficiency in low-contrast trap images, where animals are partially obscured, in shadow or nighttime conditions, using image pre-processing, noise reduction, and brightness balancing techniques. 
\item  As further conservation planning, this thesis aims to integrate a system to keep the conservation team updated on the descriptive behavior of the animals by leveraging Multimodal Large Language Models.
\end{enumerate} 
\hspace{0.3cm}

\section{Report Organization}
The organization of this thesis is structured as follows:\\[0.2cm]
\textbf{Chapter 1: Introduction}\\
\indent This chapter contains the background, research problem identification, hypothesis, and research objectives of this study.\\[0.2cm]
\textbf{Chapter 2: Literature Review}\\
\indent This chapter features a review of research papers relevant to this study, providing an overall analysis of existing studies and their connection to the research objective.\\[0.2cm]
\textbf{Chapter 3: Methodology}\\
\indent This chapter presents the comprehensive methodology including i) the proposed Swin + Bi-FPN + FRCNN architecture, ii) data collection and preprocessing procedures, iii) backbone networks and feature pyramid overview, and iv) multimodal large language model overview for the integrated wildlife conservation framework. Furthermore, the chapter discusses the deployment specifications of the object detection architectures comprising F-RCNN with ResNet50, ViT+ F-RCNN,and the introduced model Swin+Bi-FPN+F-RCNN along with the implementation specifications of the multimodal large language models\\[0.2cm]
\textbf{Chapter 4: Results and Discussion}\\
\indent This chapter presents a detailed comparative analysis of the object detection architectures comprising F-RCNN with ResNet50, ViT+F-RCNN and the proposed model with Swin transformer+Bi-FPN+F-RCNN accompanied by a thorough assessment of the multimodal large language models using NLP metrics and LLM as a judge and a discussion on the potential future research scope and limitations of the existing work.
\\[0.2cm]
\textbf{Chapter 5: Conclusion}\\
\indent This chapter summarizes the findings regarding the proposed transformer-based object detection with integrated MLLM framework and discusses the potential impact of the implementation of this comprehensive system in wildlife preservation management.\\[0.2cm]
\nomenclature{$AI$}{Artificial Intelligence}
\nomenclature{$CNN$}{Convolutional Neural Network}
\nomenclature{$FRCNN$}{Faster Region-based Convolutional Neural Network}
\nomenclature{$YOLO$}{You Only Look Once}
\nomenclature{$MLLM$}{Multimodal Large Language Model}
\nomenclature{$Swin$}{Shifted Window}
\nomenclature{$LLM$}{Large Language Model}
\nomenclature{$RPN$}{Region Proposal Network}
\nomenclature{$ROI$}{Region of Interest}
\nomenclature{$FPN$}{Feature Pyramid Network}
\nomenclature{$Bi\text{-}FPN$}{Bidirectional Feature Pyramid Network}
\nomenclature{$mAP$}{Mean Average Precision}
\nomenclature{$ViT$}{Vision Transformer}
\nomenclature{$BERT$}{Bidirectional Encoder Representations from Transformers}
\nomenclature{$SBERT$}{Sentence-Bidirectional Encoder Representations from Transformers}
\nomenclature{$LlaVa$}{Large Language and Vision Assistant}
\nomenclature{$LlaMa$}{Large Language Model Meta AI}
\nomenclature{$IDEFICS$}{Image-aware Decoder Enhanced à la Flamingo with Interleaved Cross-attentionS}
\nomenclature{$PANet$}{Path Aggregation Network}
\nomenclature{$BLIP$}{Bootstrapped Language-Image Pretraining}
\nomenclature{$GPT$}{Generative Pretrained Transformer}
\nomenclature{$NLP$}{Natural Language Processing}
\nomenclature{$ResNet$}{Residual Network}

%% file: chapters/chapter_2.tex
Bangladesh is a country of diverse ecosystems with a wide range of biodiversity and home to nearly half of all the carnivore species found in the entire Indian subcontinent \cite{tbsnewsFindingFantastic}. These wildlife species suffer from deforestation, which makes it difficult for them to locate food or procreate and causes major habitat degradation. Because of that, often wild animals enter into locality leading to dangerous situations for both humans and animals. Therefore, monitoring wildlife has become very crucial for further conservation efforts.\\[0.1cm]

There are several methods that have been used for wildlife tracking and monitoring such as manual searching, gps collar tracking, harvest record or using voice recordings of the species.  These techniques have certain constraints due to poor resolution and adverse effect on animal health. Another technique is laying out sample lines to track movements of the animals which requires manpower with professional skills. Hence, governmental agencies and other organizations that work with animal conservation widely use camera traps for field-based visual data   \cite{nichols2011camera},\cite{yang2023forest}.\\[0.1cm]

Setting up camera traps to track the movements of different animals which work by sensing their motion or heat has revolutionized the ecology of wildlife and their conservation \cite{nichols2011camera}. However, this technology is vulnerable to dynamic environments, including animal crossings which leads to a large number of datasets with no presence of wildlife \cite{article}. These datasets contain hundreds and thousands of images and videos that researchers used to inspect individually in the past. However, with the advancements of machine learning, using deep learning models, have shown promising results in the identification process with less manual human assistance \cite{norouzzadeh2018automatically}.\\[0.1cm] 



\section{Related works}

Numerous deep learning methods and their combinations have been applied over time following the need for image processing and detection in various sectors. Considering compatibility, accuracy, speed, and processing power requirements, each of these models and combinations has unique results, constraints, and advantages as well as drawbacks. \\[0.2cm]

At the earlier onset of the development of CNN, Chandrakar et al. \cite{Chandrakar2020}, proposed a study where he established that CNN architectures can provide better results in animal detection in wildlife sanctuaries using webcam images. He suggested that the manual human effort of identifying animals from a large dataset can be challenging, hence using the proposed neural network to automate the process can reduce human labor while being accurate. The dataset used for this study contained a total of 48 animals and 10.8 million classifications. 75\% of the images did not contain any animal in it creating an imbalance. The proposed method implemented a two-layer network and nine different non-conventional CNN architectures, implementing one single model at a time.  The results show that every time CNN architecture provided more precise results than SVM, PCA  LDA and LBPH algorithms. Furthermore, he added that combining several architectures and fine-tuning them can offer better accuracy and help the model to learn efficiently.\\[0.1cm]

Leorna and Brinkman \cite{articleleorna}, researched the effectiveness of AI tool integration in the field of camera trap image detection aiming to compare it with manual review. The study focused on the model MegaDetector created by Microsoft AI for Earth, which is based on a faster R-CNN processing model specifically developed for wildlife detection. The researchers tried to evaluate the performance of this module in terms of image classification in complex environments, evaluating the maximum distance and minimum detection size compared to the manual human review method. The dataset was collected from a project in Arctic Alaska (USA), where the camera was set up for a long period, in a place with limited vegetation, to collect pictures in different seasons to get a diverse dataset consisting of time-lapse and motion-activated images. The same dataset was used for both human review and MegaDetector for further processing, where MegaDetector binary labeled the dataset and human reviewers divided it into specific classifications. The overall outcome was better in MegaDetector comparatively for motion-triggered images as they have more animal presence than time-lapse images. The accuracy was above 94.6\% for motion-triggered images but for time-lapse, it was lower than 61.6\%. The results of the model were different depending on different confidence thresholds, data set type, and how they were captured. The detection limitations were a key factor as the maximum distance the model could cover was very low compared to human reviewers but, the minimum detection size was notably larger on the module than the human review, making the model suitable for small object detection. To summarise, AI tools can help handle large amounts of data but it still needs human review to reduce inaccuracy and it needs further improvements to handle time-lapse images.  \\[0.1cm]

Fennell et al. \cite{fennell} investigated the integration of AI tools, specically the MegaDetector model (is a YOLOv5 object detection model trained on several hundred thousand wildlife camera trap images to detect animals, people, and vehicles) developed by Microsoft, in the processing of camera trap images, focusing on its eectiveness compared to manual classication methods. The study aimed to assess the performance of MegaDetector in detecting humans and animals in ecological research, particularly in the context of recreation ecology in British Columbia, Canada. The researchers emphasized the signicant challenge of processing large volumes of image data generated by camera traps, which often leads to bottlenecks in data analysis and decision-making for wildlife management. The dataset for the study comprised images captured by camera traps over a week, allowing the team to observe temporal variations in human and animal detections. Fennell et al. evaluated Mega Detectors accuracy, which achieved 99\% precision and 95\% recall for human detections, and 82\% precision and 92\% recall for animal detections, all at a 90\% condence threshold. The integration of MegaDetector into the workow resulted in over a 500\% increase in processing speed, signicantly reducing the time needed for manual classication by 8.4 times. Additionally, the study revealed that using an object detection model like MegaDetector can create an index of human activity, which matches nearly perfectly with results from manual classication, exhibiting only a 0.45\% mean dierence in human detection estimates across site-weeks. The ndings suggest that such AI-driven tools can markedly enhance the eciency of wildlife monitoring without compromising the accuracy of data, while still necessitating human oversight for species identication. Overall, the research posits that employing MegaDetector allows ecologists to keep a "human in the loop," ensuring quality control while capitalizing on the speed of machine learning. The study highlights the potential of AI models to bridge the gap between data collection and actionable insights for conservation eorts, particularly underlining the importance of timely data analyses in supporting eective wildlife management decisions. This approach not only accelerates processing but also facilitates the anonymization of human images, thereby addressing ethical concerns in wildlife monitoring. The need for continuous improvements in AI capabilities was also acknowledged, particularly for the identication of species in variable environmental conditions.\\[0.1cm]

In another peer reviewed research article \cite{ani13091526}, the researchers aimed to build a framework the can classify reptiles and amphibians (primarily worked with frogs/toads, lizards and snakes) within their groups from a large number of camera trap images. The researchers acknowledged that even though there are several works regarding mammal classification, very few researchers have dived into working with reptile trap images. The dataset for this study was collected from five counties of Texas, USA by setting up trap cameras and contains highly imbalanced classes. For this study, researchers developed and compared three CNN models: self trained Custom CNN (CNN-1) and two transfer learning frameworks with pretrained models VGG16 and ResNet50. The preprocessing was done by oversampling and for augmentation purpose data generator function was used. The results show that for multiclass classification VGG16 shows 87\% accuracy which is higher than the rest and for binary classification all the models achieve high accuracy. The results confirm that pre trained models with fine tuning are better for categorizing species. This research found that self trained model like CNN show less tolerance towards augmentation in case of data deficiency and also acknowledges the overfitting tendency of deeper neural networks such as ResNet50 due to vanishing gradient problem. Also research findings show that night-vision samples provide better view for the target object due to blurred background however small reptiles and camouflaged reptiles are likely to be miscategorized due to complex feature extraction. The research was limited by small datasize and model’s ability to generalise to new species and location providing further research opportunities.\\[0.1cm]

According to Gulrajani and Lopez-Paz \cite{gulrajani2020searchlostdomaingeneralization}, domain generalization refers to the apparent problem of training models that can perform well on unseen target domains without having access to any samples from those domains during training. In their paper, the authors systematically benchmark a wide range of state-of-the-art domain generalization algorithms across four widely used datasets, PACS, VLCS, OfficeHome and TerraIncognita. They highlight that despite the popularity of various domain-specific adaptation strategies, a well tuned Empirical Risk Minimization model consistently matches or outperforms many complex domain generalization methods.In the absence of real-world target data, simulating domain shifts such as lighting changes, occlusions, resolution reduction, and image noise in both training and testing stages offers a valuable and effective strategy for examining a model’s generalization performance under varying conditions. This approach enables the researchers to approximate realistic deployment conditions, especially in contexts where collecting representative data is rather difficult. The authors advocate for carefully controlled synthetic test setups that reflect the expected challenges of the target domain, emphasizing that evaluation under simulated shifts can be a rather effective stand in when true target data is unavailable. Their work underscores the importance of transparent experimental design, robust baselines and also reproducibility in domain generalization research.\\[0.1cm]

Tan et al. \cite{ani12151976} conducted an experimental study to evaluate and compare the performance of of different deep learning object detection models for identifying wildlife in camera trap images. The paper addressed the challenge of efficiently processing large number of camera trap images by using artificial intelligence to automate the identification process in image and videos. The authors constructed a NTLNP dataset which contains images of 15 wild animals and 2 domestic animals from camera traps in the Northeast Tiger and Leopard National Park. The images were labeled in pascal VOC format. In this study, the authors evaluated three object detection architectures: YOLOv5, Cascade R-CNN with HRNet32 and FCOS with ResNet50 and ResNet101. The model performance was evaluated by training the models on day and night data separately versus together. In all the models, the YOLOv5 performed best overall which helps conclude that anchor-based one-stage models outperform both anchor-based two stage and anchor-free one-stage models which is inconsistent with the idea that deeper neural networks provide better results. Furthermore, the results also show that higher threshold might not improve better accuracy but lower threshold tend to provide more false positives. The research further shows that day and night joint training is more effective. However, the models struggled with small animal detection due to faster movement, poor image quality and the background informations also affected the model performance significantly further suggesting the need for more diverse dataset. Due to the limitation of the experimental environment, this study could not compare parameters such a running time of the models.\\[0.1cm]

Nguyen et al. \cite{Nguyen2024} started 'SAWIT (Small-Sized Animal Wild Image Dataset)' in a bid to fill a gap in samples' annotation for small animals, with important roles in habitats but difficult to detect with an elusive nature. Composed of 34,434 images and 34,820 expert-selected bounding boxes for seven classes: frogs, lizards, birds, small mammals, big mammals, spiders, and scorpions, the dataset is captured over seven months with camera traps in Victoria, Australia, in actual ecological scenarios of occlusions, motion blur, and vegetative cover. Camera traps, with collaboration between Deakin University, Arthur Rylah Institute, and citizen volunteers, operated with solar-powered batteries and motion detect algorithms for real-time day-and-night observations. The dataset reflects wildlife detectability complexity such as occlusions, atmospheric complications such as condensation, and animals blending with flora, and, therefore, a successful benchmark for wildlife studies. YOLOv5 and Faster RCNN performance in the dataset was benchmarked and displayed accuracy and computational efficiency in a compromise with each other. YOLOv5l performed best in terms of best mAP (62.6\%) and real-time performance (83 fps), and, therefore, for high-speed and accuracy requirements. Faster RCNN with HRNet backbone performed best in target detection when in motion, or for small animals, but at a relatively slow pace. In terms of accuracy and efficiency, both faltered for species such as scorpions and lizards in an environment, simply because of an indistinguishable resemblance with its environment. Re-emphasizes, yet again, the difficulty in distinguishing animals in a crowded environment and future work that could maximize accuracy with an integration with a temporal axis. The contribution constitutes a useful tool for biodiversity conservation and wildlife tracking and re-emphasizes the necessity for sophisticated detection techniques for a challenge with current state-of-the-art object detection algorithms. \\[0.1cm]

Another study \cite{inbookyolov7}, proposed a different algorithm of YOLOv7 integrated with the SGD optimization method that can process real-time videos with more accuracy and efficiency than the previous versions of YOLO as well as the R-CNN, Faster R-CNN, and SSD. The researchers collected data sets from the Wild Animal Computer Vision project and the Animal Image Dataset. The integration of Stochastic Gradient Descent (SGD) with momentum optimized the performance of YOLOv7 by reducing error and helping the model to reach to the solution faster making the model's training process efficient. Weight decay and dropout layers were implemented to improve model generalization ability and to handle the overfitting issues. Moreover, to improve learning rate performance during training, exponential decay or step decay was implemented. In comparison with the other modules, the proposed model performed notably better with 90.3\% on mAP(Mean Average Precision), 91.5\% recall, 94.1\% precision rate, and better fps rate. Even though the model provided remarkable improvements in detecting objects in complex environments with better accuracy, the authors mentioned researching further to refine the model for better optimization and to use a larger dataset with diversity for facilitating its use in wildlife conservation efforts. Also, a larger dataset consisting of diverse species can affect the performance of this proposed model because of low visibility, complex environment, or data imbalance. Hence further research to improve this was suggested by the authors. \\[0.1cm]

A research work \cite{article}, on animal detection using camera trap images as dataset and ResNet50 (Residual Neural Networks) architecture, demonstrated how the efficiency increases in automated species classification. They used a dataset of 2269 images for training and 112 images for testing the model which was collected from Tanzania’s Serengeti National Park. The dataset was divided into two categories which were “Blank” and “Non-Blank” to indicate the presence of species. Additionally, to improve the loss function “Categorical Class Entropy” was used, and for better speed and optimization during training, Adam optimizer was implemented. Furthermore, the authors evaluated their proposed model's outcome and compared it with YOLOv5, and InceptionV3. ResNet-50 with its feature-based detection process, significantly improved the overall accuracy by 94.64\% whereas the other two models had an overall accuracy of 93.18\% and 92.02\% respectively. Overall, the researchers tried to demonstrate the robustness of an automated species identification process which can mitigate manual human labor and will be more efficient than the existing models. However, the model struggled to work with some of the image data because of poor lighting or unfavorable weather.\\[0.1cm]

According to Djarot Hindarto \cite{hindarto2023use}, ResNet50V2 has achieved phenomenal performance in the identification of five different species of animals: cats, dogs, cows, elephants, and pandas, with impressive training and validation accuracies of 98\% and 96\%, respectively. This model efficiently addresses the weaknesses of traditional classification methods, which are generally slow, subjective, and prone to errors, especially when discerning animals that possess similar physical features. Highlighting, Hindarto says that deep learning in the form of CNNs automates this classification task with higher accuracy and swiftness, especially under difficult conditions like poor light or vague animal appearances. Despite the great computational power and enormous, complex training dataset needed for such models, CNNs hold immense promise to bring about a paradigm shift in this field, beginning from management or tracking wildlife to identification of endangered species or ecological studies. The study notes that the performance of the ResNet50V2 model can vary depending on the species being classified, with the model performing excellently on elephants and pandas, while it faces some difficulties in correctly identifying cats and dogs due to their similar features. Further, Hindarto echoes that the performance of any deep learning model essentially depends on quality and variability; thus, developing extensive datasets for use in deep learning training is pivotal. This present work will now demonstrate how recent techniques using deep learning offer promising performance with broad applicability to this animal classification, while CNNs also become an integral part in finding practical tools that will assist conservationists in the continuous efforts toward wildlife populations monitoring and protection. \\[0.1cm]

Liu et al. \cite{article22} researched to solve the problem of small dataset size by incorporating temporal metadata with the camera trap images. This research shows that fusing images with metadata can help solve the problem of needing large datasets for model accuracy. The data used for this paper was taken from Camdeboo dataset, which includes camera trap images and corresponding metadata of various wildlife species. The methodology includes image feature extraction using SE-ResNet50 architecture which combines ResNet50 with SE attention module, temporal feature extraction was done by residual MLP network using the Cyclical encoded data and finally the features were fusioned using dynamic MLP Module. The results show an accuracy of approximately 93.10\% in wildlife recognition which is higher compared to other CNN models (ResNet50, VGG19, ShuffleNetV2-2.0x, MobileNetV3-L, and ConvNeXt-B). This research introduces a novel approach to wildlifer recognition as incorporating metadata can leverage animal activity pattern. However, for rare and elusive species, the model still declined due to limited training data and in case of wider range of ecological contexts and datasets, the model needs to be further evaluated. \\[0.1cm]

Roy et al. \cite{articleroy} suggested an improved model called WilDect-YOLO based on YOLOv4, refining the limitations by enhancing the ability to extract finer details from the trap images. The research shows extensive result comparison which denotes the lackings of existing models in terms of precision, accuracy in extracting unique details, and real-time detection for wildlife conservation. CSPX1-n and DenseNet were integrated into the backbone- CSPDarknet-53, to improve the accuracy of feature extraction and to handle the overfitting issue while maintaining computational efficiency respectively. Also, SPP blocks and PANet were integrated into the CSPX2-n for further precision. Initially, the datasets were a collection of 1600 high-resolution images and were 16000 after data augmentation. This module outperformed the existing algorithms which were Faster R-CNN, SSD, RetinaNet, Mask R-CNN, YOLOv3, YOLOv4, and Dense-YOLOv4 by 97.87\% F1 score, 96.89\% mAP, and with better fps and real-time detection ability overall. Although the model is significantly improved in various parameters than others, it is initially trained for limited species which should be redefined making the dataset more diverse for further wildlife conservation. In addition, the initial dataset mostly contained larger animals, minimizing the accuracy of the detection rate for smaller ones, hence there is scope for further development. \\[0.1cm]   

Liu et al. \cite{liu2021swintransformerhierarchicalvision} proposed an advanced vision transformer called Swin Transformer, which can be used as a backbone in computer vision. The traditional Vision Transformer which has been widely used in processing textual data, struggles with visual data which varies in size and resolution. Vision transformer leverages a fixed patch size and computes self attention across all the other patches and that leads to higher computational complexity which is not suitable for tasks that demand detailed processing. The proposed “Swin transformer" lowers the computational complexity to linear by introducing a window based self-attention mechanism. The image is divided into 4*4 patches and as the layers progress the patches are merged together with their neighboring patches. Instead of calculating the global attention, this approach limits the calculations within the neighbouring patches in a window which helps to reduce the complexity. Furthermore, there is a shifted window mechanism introduced here, that is the windows are shifted by M/2 patches, in successive  transformer blocks alternatively, which allows patches from different non-overlapping windows to connect and this process helps to get global context while maintaining linear complexity. The shifted window mechanism creates smaller windows which are managed efficiently by the cyclic shifting mechanism. The authors constructed four variants of this transformer with different levels of complexity based on parameters. It outperformed the previous vision transformer and other existing CNN’s in computer vision tasks such as object detection, image classification and semantic segmentation in terms of precision, accuracy and efficiency. The swin transformer efficiently connected different domains and perfectly incorporated with the existing frameworks.\\[0.1cm]

In another article \cite{yang2023forest}, the authors worked on improving wildlife detection accuracy using trap images by introducing an improved algorithm based on YOLOv5s and Swin transformer. It addresses the high detection error and omission rate due to low background contrast and serious occlusion, data imbalance etc. Dataset for this research article were divided into two sections, dataset 1 includes images of five rare wildlife species from Hunan Hupingshan National Nature Reserve in China of over the 5 years and dataset 2 includes previous five categories of wildlife, as well as screened subset of the 2019 iWildCam Wildlife Identification public data set filmed in North America which is an international competition dataset for wildlife recognition to generalise and complete the model. For the dataset preprocessing, annotation was performed manually using the open-source tool Labeling and data enhancement and augmentation were performed using computer vision techniques to enrich the data set, including rotated image adjustment, Gaussian blur noise, and image fusion(Cutout and Cutmix method). The researchers used YOLOv5 as the backbone and Swin transformer as the neck, added SENet channel attention mechanism, and used improved loss function: DIOU\_loss and adaptive class suppression loss. The results of the above improved algorithm achieved 89.4\% mAP, improving accuracy by 16.8\% compared to original YOLOv5s and outperformed other models like YOLOv3, Faster R-CNN and RetinaNet. While this research show an improved detection of small animals and overlapping animals this model had to trade off between model complexity and inference speed and the GPU optimization of the transformer models need improvement. \\[0.1cm]

Buslaev et al. \cite{albumentations} developed a python library named “Albumentations”, which provides numerous augmentation techniques with higher efficiency and performance than the others. The existing libraries used for augmentation had very few methods to provide which was not sufficient for the tasks that needed complex transformations such as adverse weather conditions or motion blur effect. Also with the advancement of GPU’s, the existing frameworks and libraries were facing issues with slower processing CPU’s creating bottleneck in the training process. Hence, the authors focused on improving the efficiency, adaptiveness and optimization in the “Albumentations” library to further speed up the augmentation process in computer vision. This library utilizes several low-level libraries and combines them for optimized processing, implements libraries such as numpy instead of loops and also uses “uint8” format for image processing that helps to lower memory requirements. Furthermore, it has the ability to transform bounding boxes and masks with appropriate annotations, hence it is suitable for all kinds of computer vision tasks. The paper further discussed a comparative study on the performance of “Albumentations” which showcased that it outperformed the existing libraries like “Augmentor”, “imgaug” in terms of speed in almost all types of transformations. The researchers also implemented heavy augmentation on the “Inria Aerial Image Labelling Dataset” using “Albumentations” which showed a massive improvement on the models performance while mitigating CPU bottleneck. In future, the authors wanted to work on GPU based augmentation to further enhance the training process and also wanted to work on the library so that it can be used for three dimensional transformations as well. \\[0.1cm]

Wang et al. \cite{wang2024visiongptllmassistedrealtimeanomaly} presents a real-time, accessible visual navigation system aimed at assisting visually impaired and blind individuals in complex environments. The authors leverage a combination of open-world object detection models, specifically YOLO-World, and large language models (LLMs), such as GPT-3.5 and GPT-4, integrated through carefully engineered prompts to enhance scene understanding and hazard detection. The system processes live camera feeds to identify potential obstacles and anomalies, generating descriptive audio alerts that emphasize safety-critical information. Wang et al. highlight the limitations of traditional rule-based detectors and emphasize the benefits of prompt engineering in adapting LLM responses to dynamic scenarios. Their approach involves optimizing detection accuracy and computational efficiency to enable deployment on mobile devices, ensuring low latency for real-time feedback. They address challenges such as balancing detection precision with processing speed, particularly in resource-constrained environments, through strategies like multi-frame processing and parallel operation of different modules.It processes camera frames continuously, achieving low latency (~60 milliseconds) on mobile devices, and employs multi-frame analysis to improve accuracy and stability. The results demonstrate high detection accuracy, effective hazard alerts, and adaptability across different hardware, significantly enhancing safety and independence for users. Through extensive evaluation across multiple platforms, Wang et al. demonstrated high accuracy and low latency, confirming the system’s viability for real-world use. They also discuss the potential for further improvements in prompt design and model architecture to better handle complex and unforeseen scenarios. \\ [0.1cm]

 Tian et al. \cite{TIAN2024101875}, introduce KED (Knowledge Enhanced ECG Diagnosis), a foundation model that represents a significant advancement in automated electrocardiogram interpretation. The study therefore addresses limitations of traditional ECG diagnostic models, which happen to be typically trained on narrow datasets and struggle to generalize across diverse patient populations as well as clinical settings. To overcome these challenges, the authors design a signal language architecture that integrates raw ECG signals with medical knowledge using large language models. The fusion is apparently achieved through Augmented Contrastive Learning or AugCL, which aligns signal, text and structured label representations, thereby enhancing semantic understanding of cardiac conditions. A central component happens to be the Label Query Network, which allows natural language queries about potential abnormalities. The model returns diagnostic probabilities and interpretable explanations aided by Grad-CAM as well as GPT-based text generation. Trained on 800,000 ECGs, KED was evaluated on five external datasets spanning different regions and patient demographics. Remarkably, it demonstrated strong zero-shot performance, successfully identifying conditions not exposed to during training, such as ST segment depression as well as supraventricular tachycardia. The model also exhibited superior accuracy to experienced cardiologists across clinical benchmarks. These results therefore highlight KED's robust generalization and potential for real-world deployment, particularly in low-resource healthcare environments where rapid ECG interpretation is rather critical. \\[0.1cm]

According to Zheng et al. \cite{zheng2023judgingllmasajudgemtbenchchatbot}, traditional LLM benchmarks like MMLU, ARC etc. inadequately measure what users actually value in conversational AI, creating a sort of misalignment between benchmark scores and real user satisfaction. To address the gap, the authors developed two human preference benchmarks, MT-Bench, featuring 80 multi turn conversations across eight categories and Chatbot Arena, a crowdsourced platform collecting over 30,000 user votes comparing anonymous chatbots.The researchers systematically explore using strong LLMs like GPT 4,Claude as well as, GPT 3.5, as automated judges to evaluate other models through pairwise comparison, single answer scoring and reference guided grading. While this "LLM-as-a-Judge" approach offers scalable, cost-effective evaluation with more explainable reasoning, Zheng et al. identify significant biases including position bias (favoring first presented answers), verbosity bias (preferring longer responses) and limited reasoning ability leading to misgrading. To mitigate these issues, the authors propose swapping answer positions, few-shot prompting with examples and also chain-of-thought reasoning for complex tasks. Their empirical results show GPT-4 judges achieve over an 80\% agreement with human evaluators, matching human to human agreement levels. This demonstrates the ultimate fact that LLM-as-a-Judge can reliably approximate human preferences at scale, providing a rather practical solution for evaluating conversational AI systems in such ways that traditional capability based benchmarks cannot at any instance, capture. \\[0.2cm]

Based on a review of recent research articles, some opportunities for further investigations have appeared. Although deep learning models have demonstrated competitive performance, challenges remain in accurately detecting small animals in low-contrast trap images because of inadequate feature extraction \cite{ani12151976}, \cite{ani13091526}. While some studies have proposed attention-based mechanisms within object detection models to enhance performance, the results are still preliminary and subject to further validation regarding small animal detection in complex or severely occluded images \cite{yang2023forest}.\\[0.05cm]
Traditional object detection research solely focuses on detecting the object. While it is crucial, the detection results can be insufficient in terms of wildlife conservation efforts. Further information about the animal’s behavior can provide important cues about any anomaly observed. Moreover, traditional approaches primarily rely on supervised learning from annotated training data, which can be expensive in terms of all types of resources. In the modern research terrain, a few emerging works have explored the potential of large language models (LLMs) and vision-language models, leveraging their extensive pretraining and open-source capabilities to enable zero-shot detection and visual semantic extractions with promising accuracy \cite{TIAN2024101875}, which is worth exploring for our thesis, as it provides the opportunity of detecting rare animals without having to rely on trained data only and also provides important data for conservationists.\\[0.05cm]
There is also a notable scarcity of research focused on the wildlife of Bangladesh, especially in the context of trap images. The lack of annotated trap image datasets presents an additional challenge \cite{ani12151976}, \cite{ani13091526}, \cite{inbookyolov7}. Simulating trap image like conditions to address this data gap remains a critical area for image vision research \cite{gulrajani2020searchlostdomaingeneralization}.
Hence, this work aims to bridge these gaps by exploring the integration of attention-based models for wildlife detection in trap images while leveraging the multimodal large language model’s zero-shot detection capability and visual semantic extraction, with a specific focus on ecosystems such as those in Bangladesh.
\\[0.1cm]

%% file: chapters/chapter_3.tex
\begin{figure}[H]
\centering
\includegraphics[width=\textwidth]{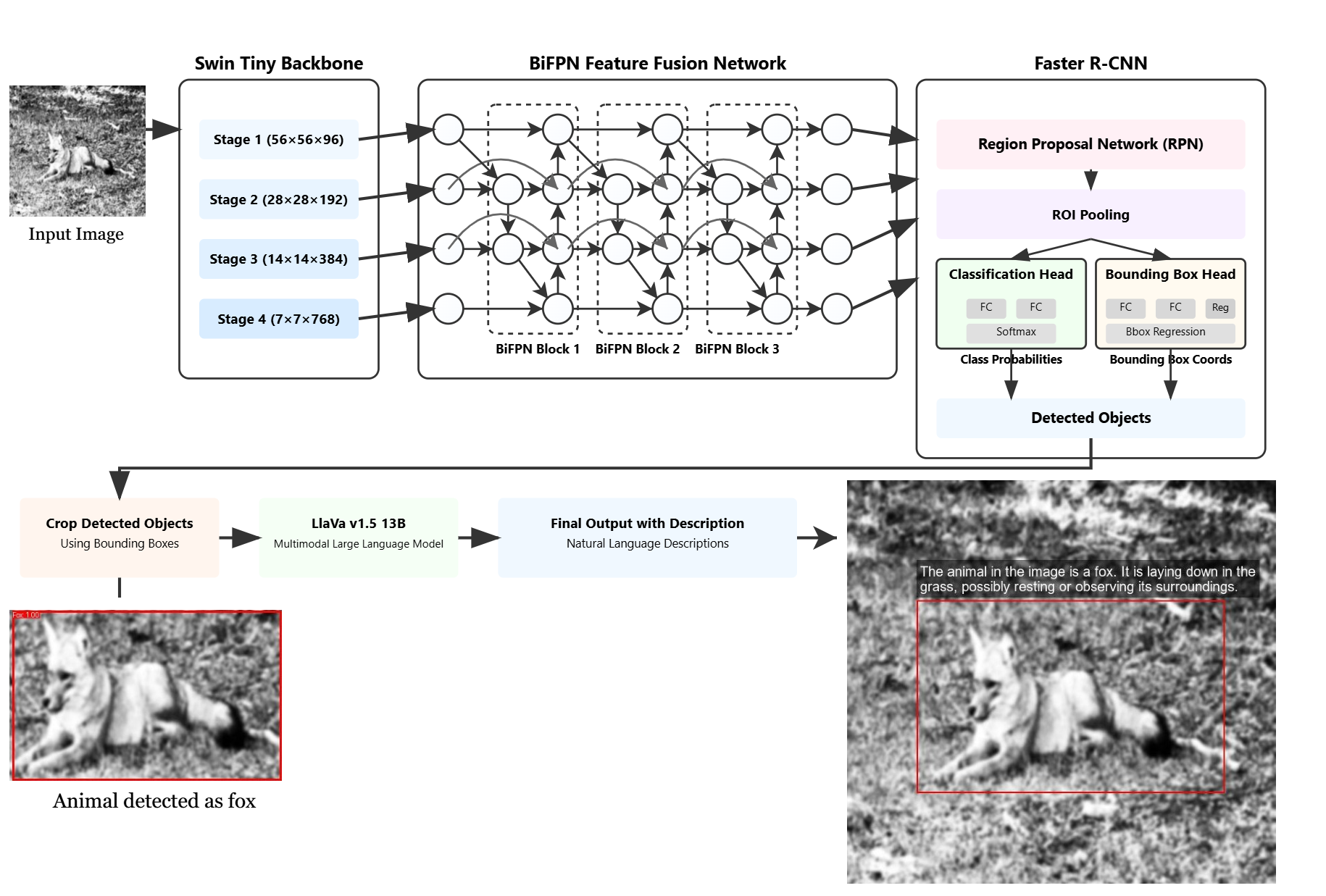}
\caption{Proposed Architecture: The model integrates a Swin Transformer and Bi-FPN within the Faster R-CNN backbone, complemented by LLaVA v1.5 (13B) for vision semantic extraction} \label{final}
\end{figure}

\section{Proposed Architecture}
As illustrated in the figure \ref{final}, our proposed architecture involves a Swin transformer backbone integrated with a three-layer Bi-Directional Feature Pyramid Network within a Faster-RCNN detection framework. The swin transformer, along with Bi-FPN, works as a feature extraction module for the Faster RCNN detection network. Subsequently, the Region Proposal Network (RPN) and ROI Pooling refine the candidate regions, enabling the network to make predictions based on a confidence threshold. \\[0.05cm]
The Swin transformer module divides the images into patches, computes self-attention within local patches, and alternatively uses a shifted window mechanism. This allows the module to extract better features with low computational overhead. Feature maps from all four stages of the Swin Transformer are extracted and fused using a Bi-directional Feature Pyramid Network (Bi-FPN), which aggregates features through a weighted top-down and bottom-up pathway. The Bi-FPN allows low-level semantic features to be merged with high-level spatial information, allowing enhanced feature maps for the Faster RCNN network. After the features are extracted using the Swin-BiFPN backbone, these are passed to the Faster CNN network for detection.\\[0.05cm]
After the detection, each detected bounding box is cropped and passed to the pretrained LLaVa-v1.5 13B model, which generates the descriptive output. The overall system consists of two crucial tasks: object detection and visual semantics extraction, allowing robust animal detection and zero-shot detection capabilities. While object detection is essential for accurate population counting, the visual semantics extraction component provides conservationists with valuable insights into potential behavioral abnormalities in animals.

\section{Data Collection and Preprocessing}

\subsection{Dataset: Animal Detection Images Dataset}
\begin{figure}[H]
\centering
\includegraphics[width=\textwidth]{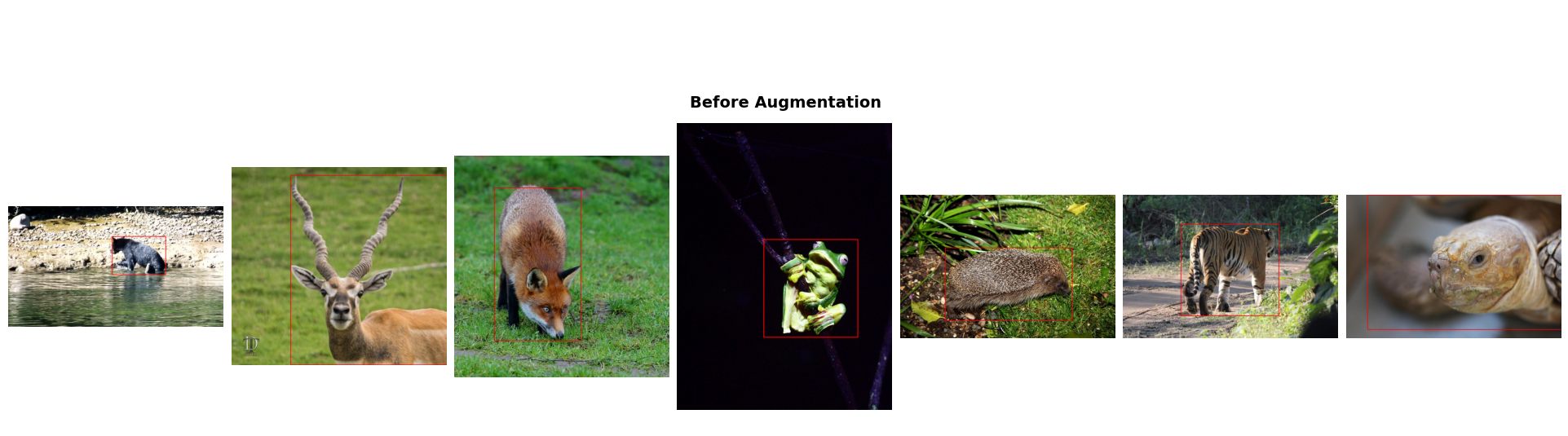}
\caption{Images from raw dataset} \label{raw}
\end{figure}

The dataset used for this study is titled “Animal Detection Images Dataset” collected from Kaggle which consists of images extracted from “Google Open Images V6+”. Open Images is the largest dataset composed of millions of images accumulated from various sources with annotations, bounding boxes and labels that facilitate different computer vision tasks such as object detection, image classification or visual relationship detection. The “Animal Detection Images Dataset” is a subset of “Google Open Images” focusing on animals and it is composed of 80 different classes with 29071 images in total. Based on the nature and variations of the wildlife in Bangladesh, 7 distinct classes are chosen among these 80 classes, to further support the conservation efforts for these animals. The selected classes are Tiger, Bear, Deer, Fox, Frog, Hedgehog, and Turtle which comprises a total of 2001 images. Manual annotation has been done on the dataset to ensure precise labeling of the images and correctly classify them to the designated classes. In figure \ref{raw}, some of the images from the selected dataset has been showcased. \\[0.1cm]

The dataset is highly imbalanced as one-third of the total images belong to the “Frog” class whereas the “Turtle” class has only 29 images. The images mostly exhibits similar lighting conditions with different backgrounds in some of them. Also, the majority of the images are taken during daytime and in suitable environmental conditions which is not useful for the research as the aim of this study is to develop a model that can detect animals from camera trap images. Hence, data augmentation is applied to introduce challenges in the dataset that can further help the models to perform better for different environmental or weather conditions.  \\[0.1cm]

\begin{figure}[H]
\centering
\includegraphics[width=\textwidth]{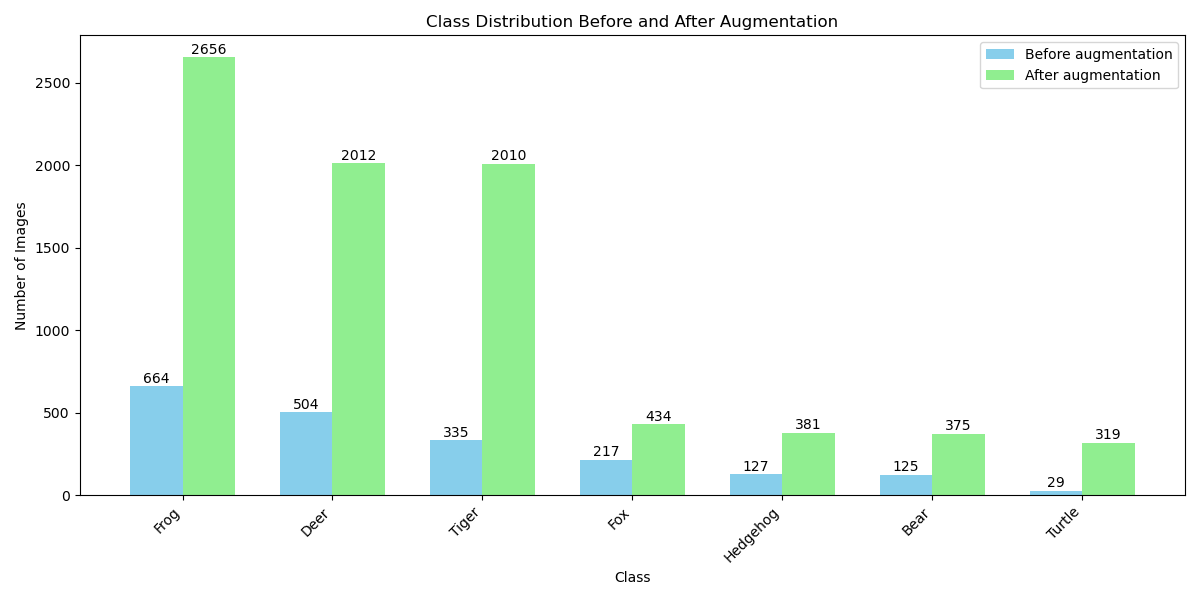}
\caption{Graph showing class distribution before and after augmentation} \label{distribution}
\end{figure}

In the figure \ref{distribution}, the number of images for each category before and after augmentation is illustrated. It is evident that images for each class have increased; however, it has increased in a reasonable range, ensuring model generalization without introducing risks of overfitting.
\subsection{Dataset Annotation and Augmentation}
The original images of the dataset are labeled manually using the open-source tool RoboFlow. Since the dataset did not have uniform pixel size, the images are scaled to 720 x 720 size using the Albumentation library. The images of the animal classes had data imbalance and did not have trap image-like challenges, so to handle class imbalance and introduce challenges oversampling and synthesized images are used by augmentation of the original images. The oversampling factor was carefully determined using an equation derived to avoid overfitting issues. The factor was:\\[0.1cm]

\[
n = 
\begin{cases}
\left\lceil \dfrac{N_{\text{total}}}{N_{\text{class}}} \right\rceil & \text{if } \dfrac{N_{\text{total}}}{N_{\text{class}}} \leq 6 \\
\left\lceil \dfrac{N_{\text{total}}}{N_{\text{class}}} \times 0.15 \right\rceil & \text{otherwise}
\end{cases}
\]

Where,  
\begin{align*}
    N_{\text{class}} & = \text{Number of images in the class} \\
    N_{\text{total}} & = \text{Total number of images in the dataset} \\
    \textit{n} & = \text{Number of augmentations for the class}
\end{align*} \\[0.1 cm]

Here, for this study, the ratio threshold is set to 6 and if the ratio of the images in a specific class exceeds this threshold, the ratio is scaled by 15\% which is suitable for the dataset used in this study to set the number of augmentations. This can be fine-tuned according to the dataset to make sure the number of augmentations does not become 1.

\begin{figure}[H]
\centering
\includegraphics[width=\textwidth]{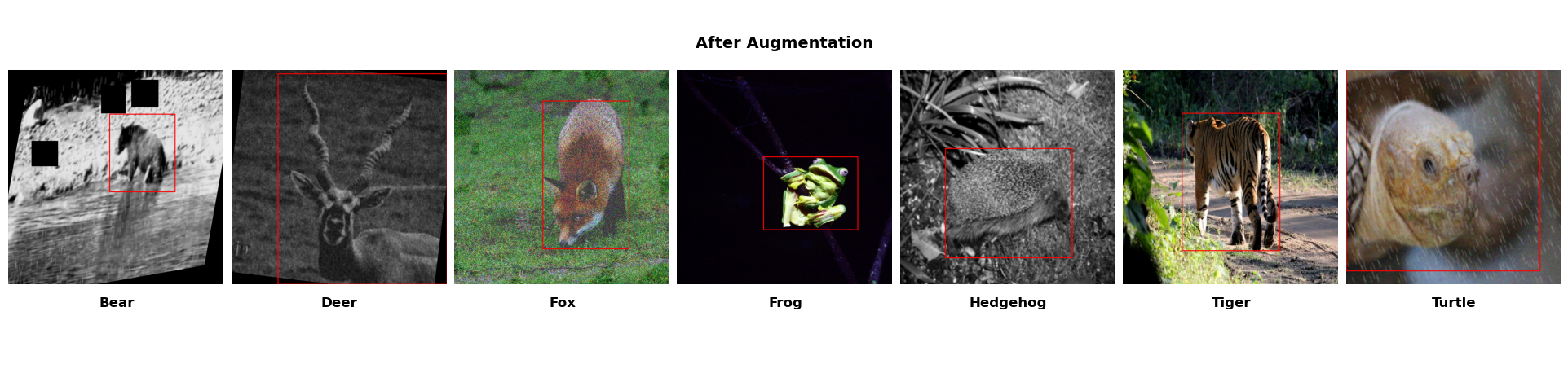}
\caption{Images after augmentation} \label{after aug}
\end{figure}

The figure illustrates the augmentation and preprocessing pipeline for the wildlife image dataset. Augmentation techniques are tailored to replicate conditions typical of camera trap images, such as motion blur, poor lighting, and occlusion. After resizing, to introduce challenges using augmentation methods, computer vision techniques are used by using the Albumentations library to introduce random rotation, horizontal flip, Gaussian Noise/Blur, Motion blur, Random Brightness Contrast, Random Grayscale, Random fog/ rain, Affine to skew the image, and lastly, CutOut method using Coarse Dropout where random rectangular regions of the image are masked to improve the robustness of the model. All of these techniques are applied based on random probability, which helped to introduce multiple challenges such as night vision-like images using random Grayscale and brightness contrast, foggy and rainy images, and images from different perspectives using skew, flip, and rotate (figure \ref{after aug}). Resizing and augmenting using the Albumentation library enabled to preserve the aspect ratio of the original image and the bounding box and the augmented images contained augmented bounding boxes. These synthesized images help enrich the dataset and improve the model's generalization by imitating natural conditions.

\begin{figure}[H]
\centering
\includegraphics[width=\textwidth]{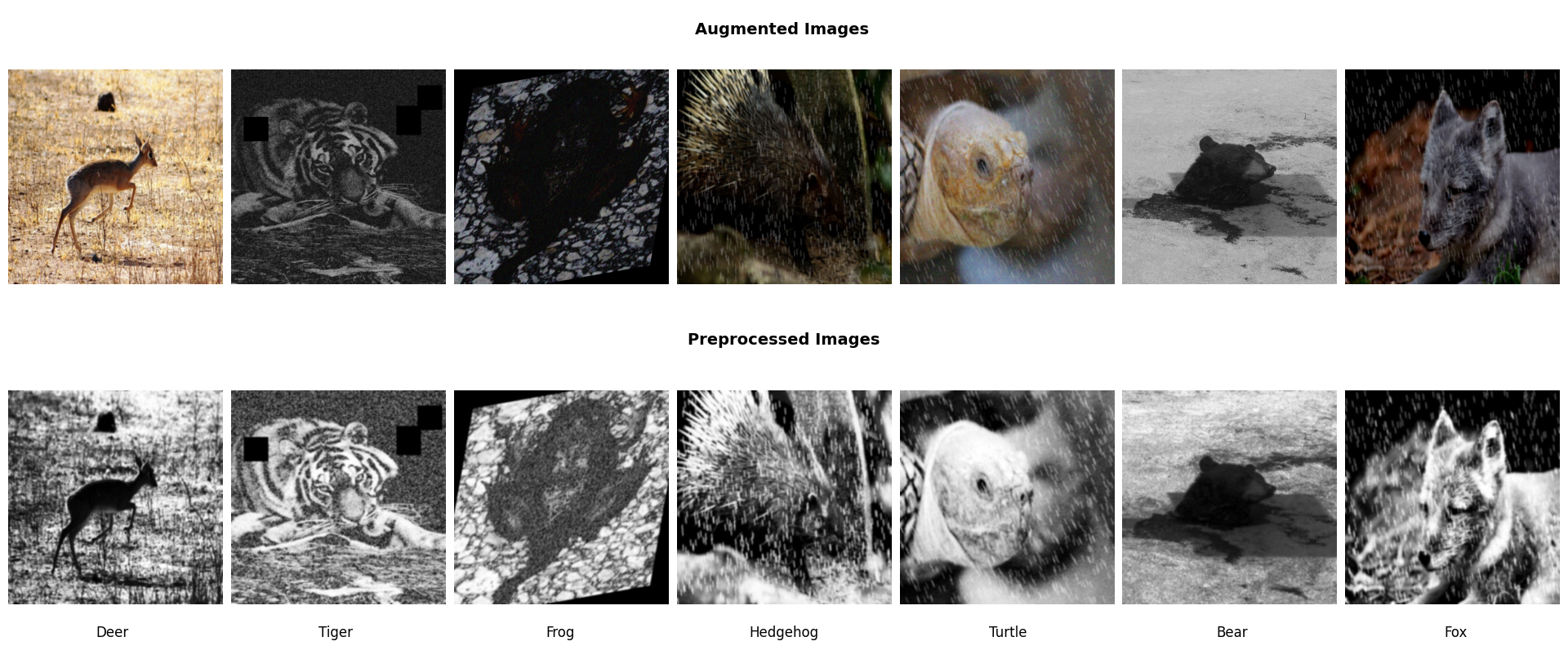}
\caption{Images after preprocessing} \label{after pre}
\end{figure}

\begin{figure}[H]
\centering
\includegraphics[width=\textwidth]{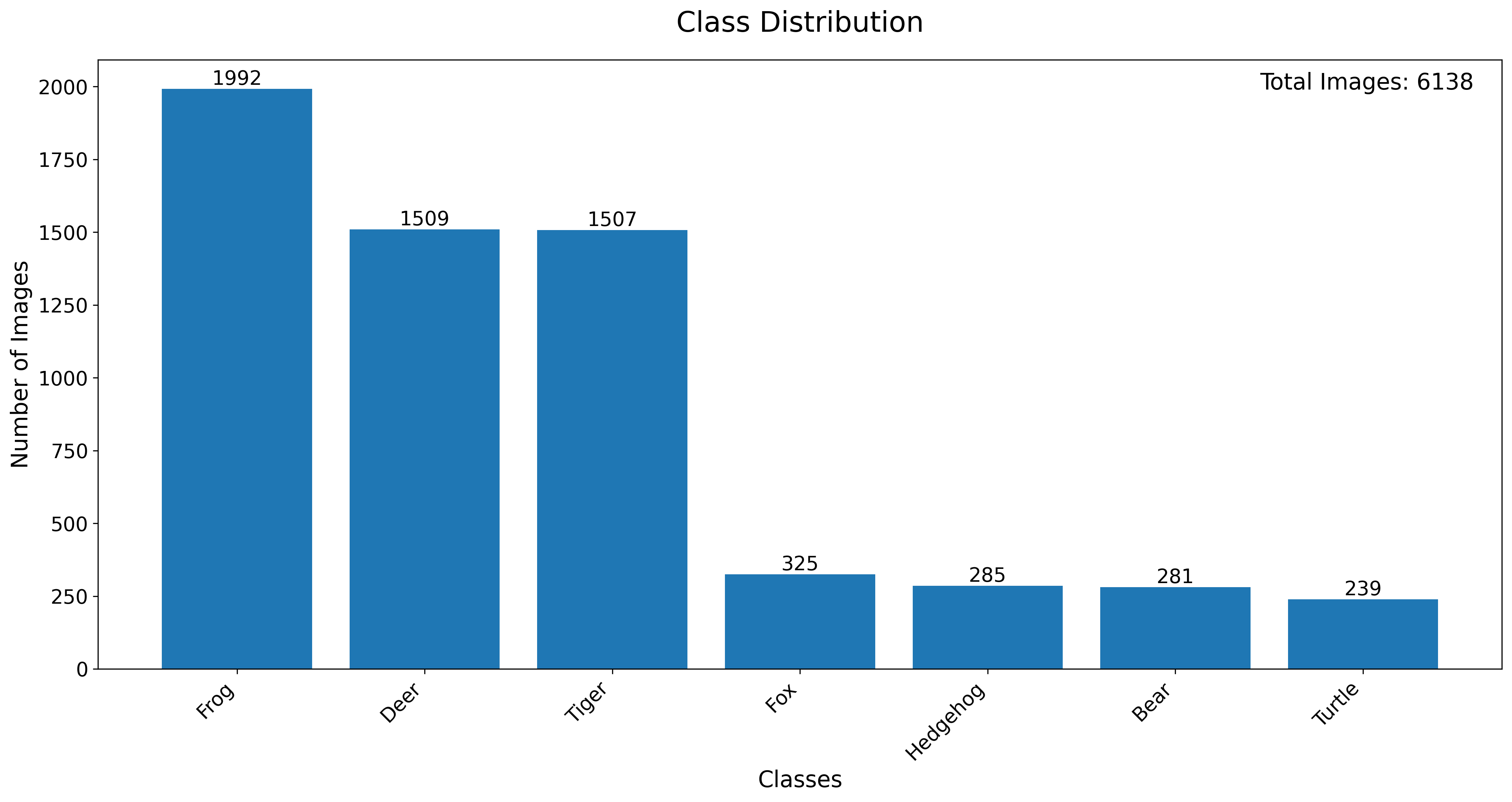}
\caption{Graph showing class distribution of training dataset} \label{train}
\end{figure}

After the augmentation process, the dataset increased to a total image of 8187. Even though our dataset was augmented using a strategically derived formula-based approach that dynamically varied the number of augmentations applied to images for each category, the risks of overfitting still persisted. Hence, to mitigate further overfitting issues, 25\% of the augmented dataset was intentionally discarded. This step ensured that the model did not learn redundant patterns or become biased toward augmented data, thus maintaining the model's generalization capability on unseen data. This resulted in 6138 images for training purposes. The figure \ref{train} demonstrates the class distribution of the final training dataset, as the 25\% discard stabilizes the dataset a bit further. \\[0.1cm]
The training dataset images are further enhanced using some more preprocessing techniques as illustrated in the figure . Normalization has been applied to normalize the pixel values to ensure consistent input distribution, Histogram Equalization to improve the contrast of the image by redistributing the intensity values of the pixels, and Gaussian blur to remove the noise and blurriness of the images.

In order to ensure unbiased model evaluation, a separate test dataset of 620 images was curated by combining samples from various publicly available datasets. This ensures the result's validity while mitigating any bias. Augmentation techniques were applied on the test dataset to introduce trap-image like challenges without oversampling. To thoroughly evaluate the model’s performance, the models were tested on both raw and augmented images, after preprocessing.\\[0.1cm]

\subsection{Reference Caption Generation for MLLM Evaluation}
To evaluate the performance of the visual semantic extraction module, a total of 700 images were selected. This included 600 manually chosen samples from the existing test dataset, along with an additional 100 randomly selected animal images from other categories not present in the training set. Those 100 images were selected from the original animal detection images dataset. The inclusion of these 100 out-of-distribution images was specifically intended to assess the zero-shot capabilities of the MLLM. After the image dataset creation, the descriptions were crafted for each image, and a textual dataset was created. For all MLLM analysis, these 700 images, along with 700 textual descriptions, were used as the ground truth reference. 

\section{CNN-based Classification and Object Detection Models}

After the train-test split stage, four state-of-the-art models: Faster R-CNN, EfficientNetV2, ZF-Net, and YOLOv11. These four models are chosen based on their efficiency to analyze how these models perform on the dataset. Following that, the models are assessed based on how accurately they can provide a result that aligns with the motive of the study. The results are evaluated to meet the requirements needed for the research. If the outcomes are acceptable, the model can be used for further classification of unseen data. Otherwise, the model is fine-tuned until it meets the requirements.

\subsection{Faster-RCNN}

Faster R-CNN is an extension of Fast-RCNN that provides more rapid results by generating and ROI pooling the Region Proposal Networks (RPN) along with the Fast-RCNN backbone. ROI pooling utilizes max pooling to extract a uniform feature map, ensuring a fixed-size representation for regions of varying dimensions. RPN works like a selective search algorithm by telling the Fast-RCNN where to look. The RPN and the convolutional layers share the computations, making the detection process faster.

\subsection{EfficientNetV2}

EfficientNetV2 is an improved version of the original EfficientNet architecture in terms of speed, accuracy and efficiency. The improvement came from three crucial optimizations which were- introducing Fused-MBConv Blocks, progressive learning strategy and using NAS - Neural architecture search. Implementation of Fused-MBConv blocks combined with MBConv blocks significantly increased the overall performance and accuracy while NAS helped to determine the best combination of these two. And the progressive learning strategy reduced the models training time to learn complex data by gradually increasing the size and regularization intensity.

\subsection{ZF-Net}

ZF-Net is another SOTA model that uses smaller convolutional filters and smaller strides to ensure better feature extraction. ZFNet utilizes 7x7 filters in the early layers to capture large-scale features and transitions to 3x3 filters in later layers for finer detail extraction. This approach ensures effective feature representation while maintaining computational efficiency.

\subsection{YOLOv11}

Yolov11 is the newest version of YOLO (you only look once), which consists of improved backbone and architecture with C3K2 block, C2PSA which is an improved attention mechanism and SPPF (Spatial Pyramid Pooling Fast). This refined architecture helps the model to detect smaller objects in a complex environment with greater accuracy by extracting critical features. Also the training procedure is much faster and optimized than the previous versions with the refined training pipeline. 

\section{Backbones for Faster-RCNN}
\subsection{ResNet 50}

Resnet50 is a prominent residual neural network (49 convolutional layers and 1 fully connected dense  layer) used in various computer vision tasks such as object detection, image classification or segmentation. This network primarily solves the “vanishing gradient” problem during the training phase in deep neural networks. When it is adapted as a backbone, it functions as a feature extractor generating hierarchical features at different stages which are used for regional proposals and further classifications. 

\subsection{Vision Transformer(ViT)}

Vision transformer is a renowned transformer architecture that has been used widely in the vision and language domain. In computer vision tasks, ViT interprets image inputs as a series of patches with a fixed size (16*16) for further processing. The patches are flattened and transformed into 1 dimensional vector and positional embeddings are incorporated to indicate the initial placements of the patches in the image. Unlike the CNN architecture’s, ViT computes self attention among all the patches which helps to gain global context from the very beginning and can detect dependencies among far-reaching patches. It has a CLS token which is attached to the patch embeddings at the beginning and is used to provide the final classification output. While implemented as a backbone, ViT encounters some difficulties such as- the computational complexity increases because of the global self-attention mechanism and it is comparatively less versatile in terms of input dimensions, which requires complex modifications to function with existing frameworks and to utilize it in object detection. Besides that, ViT can generate strong feature representations from images consisting of complex environments which is beneficial in computer vision. 

\subsection{Swin Transformer}

\begin{figure}[H]
\centering
\includegraphics[width=\textwidth]{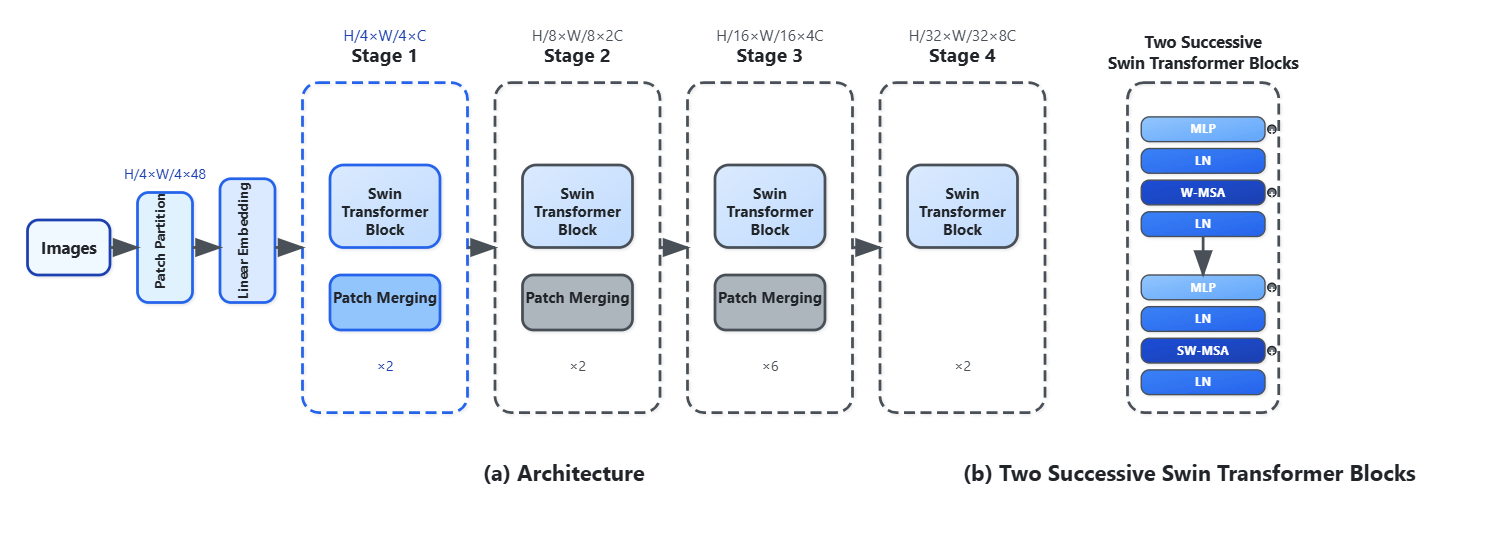}
\caption{Swin Transformer Architecture} \label{swin}
\end{figure}

Swin Transformer is an improved transformer architecture that fuses the benefits of attention mechanism and multi-scale feature extraction while maintaining linear complexity in computer vision tasks. The architecture is illustrated in figure \ref{swin}. It separates the input image into non-overlapping windows (7*7 patches) and computes self-attention among the neighbouring patches within that window. The shifted window mechanism leverages interaction among the non-overlapping neighbouring windows which helps to get a global perspective while maintaining linear complexity. The swin transformer has a 4 stage hierarchical structure, each consisting of two sections that computes window based multi-head self-attention and shifted window based multi-head self-attention alternatively. Compared to ViT, swin transformer works seamlessly as a backbone for object detection tasks because of its multi-scale feature production and compatibility with existing frameworks. The features extracted from different stages can  be used by FPN directly to create feature pyramids in object detection tasks. Also, the computational cost is significantly less with the ability to extract both local and global features which is beneficial for processing complex environments. 

\section{Feature Fusion Networks}

\subsection{Feature Pyramid Network}

Feature Pyramid Network is a neural network architecture used in the object detection task. It extracts feature maps at multiple scales to detect objects of different sizes. FPN can extract hierarchical features by constructing a top-down path with lateral connections to combine features of low resolution that are rich in semantics and features of high resolution that are rich in spatial information. This improves accuracy in detecting small objects and large objects simultaneously.

\subsection{Bi-Directional Feature Pyramid Network}
Bi-Directional Feature Pyramid Network (Bi-FPN) is a weighted feature pyramid network that is efficient in multiscale feature fusion, which makes it reliable for enhanced feature extraction. It allows the network to retain spatial information and aggregate features by adjusting the weights of each input feature map accordingly. This is faster as the regular convolutions are replaced with depthwise separable convolution. The combined top-down and bottom-up parsing, along with weighted future fusion, allows better feature extraction, which further helps the classification network to produce competitive results. Furthermore, Bi-FPN allows cross-scale connections that introduce additional connections between non-adjacent feature levels that enrich information flow. Bi-FPN produces superior feature fusion at both high and low levels and also shows significantly reduced computational overhead than FPN and PANet.  

\section{Multimodal Large Language Models}

Multimodal Large Language Models (MLLM) are deep learning systems capable of understanding and generating content across multiple modalities: typically text, image, audio etc. Unlike traditional large language models that can only operate on texts, MLLMs are trained on multimodal data and can process information from various modalities seamlessly. This flexibility allows these models to be integrated with computer vision tasks such as object detection.

\subsection{LlaVa v1.5}
LLaVa-1.5 is an advanced large multimodal model which incorporates LLM with a vision encoder through a Multi-layer Perceptron (MLP) for tasks that require language and vision-related understanding such as image captioning, generating textual descriptions for images and question answering from the visual information. Vicuna is working as the foundational language model with a pre-trained vision encoder CLIP ViT-L/14. LLaVa models are capable of managing multiple images and prompts as inputs which makes it suitable for complex tasks. We have used two different versions of this model which are LLaVa-1.5-7B-hf and LLaVa-1.5-13B-hf. 

\subsubsection{LLaVa-1.5-7B-hf}LLaVa-1.5-7B-hf architecture is based on transformer with 4 bit quantization and it utilizes Flash-attention 2 which makes it more optimized. The model has a total of 7.06 billion parameters with FP16 precision.

\subsubsection{LLaVa-1.5-13B-hf}LLaVa-1.5-13B-hf is a more powerful model which is capable of advanced reasoning, generates better textual description with more detailed knowledge because of the parameter count being increased to 13 billion. This model has the same architecture as the 7B-hf version and is trained on the same dataset. The large number of parameters impacts the overall performance of this model and makes it more robust.

\subsection{LlaVa v1.6 Mistral}
LLaVa mistral is another large multimodal model belonging to the LLaVa-Next family, which is known for its prominent efficiency in generating textual description of images with enhanced reasoning capability and OCR (Object Character Recognition). The model has Mistral-7B-Instruct-v0.2 as the base language model with the pretrained vision encoder CLIP ViT-L/14 linked through the Multi-layer Perceptron (MLP) with GELU activation. It was trained on approximately 7 billion parameters with an extensive dataset consisting of 558k image-text pairs with captions, 158k GPT generated multimodal instruction-following data, 500k academic task oriented data, 50k GPT-4 and 40k shareGPT data. 

\subsection{KOSMOS 2}
KOSMOS 2 is a transformer based Multimodal Large Language Model that is trained on the grounded image-text pairs known as the GRIT dataset. This pretrained model has grounding capability, which allows it to directly output the object’s coordinates as language tokens in Markdown syntax. The data format is similar to hyperlink. The model has a total of 1.6B trainable parameters, positioning it as relatively lightweight compared to other MLLM models.
\subsection{IDEFICS}
IDEFICS (Image-aware Decoder Enhanced à la Flamingo with Interleaved Cross-attentionS) is an open source MLLM model that is a reproduction of Flamingo, which is a closed source visual language model. The model has the ability to process visual information from images and answer questions. IDEFICS builds on two pretrained parent models—CLIP ViT-H/14 (trained on LAION-2B) and LLaMA-65 B. Both are open-access, unimodal models that enable IDEFICS to bridge two modalities: text and image. The model has two variants: one with 9 billion parameters and the other one with 80 billion parameters. For our study, we have opted for the 9B parameter one.

\section{Implementation Details}
To start with, a few preprocessing was done on the images to make sure the input format was compatible with the Swin transformer backbone. The trap images were in grayscale format after augmentation which were converted into RGB format with three channels to meet the input requirement of Swin Transformer using PIL’s method. Following that, the images were resized to a standard size of 224*224 pixels to maintain a compatible format for the patch embedding process. \\[0.1cm] 

After that, the images were passed into the Swin Tiny backbone for feature extraction. There, the input images were divided into windows of non-overlapping patches with a size of 4*4 pixels, so the 224*224 input images with 3 dimensions were converted into 3136 tokens. Every single token or patch was linearly embedded into a feature vector of 96 dimensions. Swin tiny has a hierarchical structure with four stages, each consisting of two sections that computes window based multi-head self-attention and shifted window based multi-head self-attention alternatively. Each of the stages has a different number of transformer blocks which are 2,2,6 and 2 accordingly. In every stage, the patches were merged together reducing the resolution and scaling the feature dimension by a factor of 2. By leveraging window based self-attention computation combined with the shifted window mechanism, the model extracted multi-scale hierarchical features from every stage which included precise details along with global context from the trap images.\\[0.1cm]

Subsequently, these hierarchical features extracted from four stages were passed onto a 4 level BiFPN architecture, which has three layers with cross-scale connection across non-contiguous levels. The bidirectional feature network fused features from both top-down (p6->p5->p4->p3) and bottom-up (p3->p4->p5->p6) direction ensuring that every feature level had both higher level features along with lower level details. At each layer, BiFPN utilized dynamic weighted feature fusion to evaluate the impact of the features and their contributions which enhanced the overall representation in consecutive layers.\\[0.1cm] 

After that, the multi-scale feature maps were sent to the Faster R-CNN architecture for final classification. In the training phase, AdamW optimizer was implemented with a learning rate and weight decay of 0.0001. The training was conducted using a batch size of 32 over 10 epochs and a CosineAnnealingWarmRestarts scheduler was utilized to elevate the learning rate with a minimum rate of \(1 \times 10^{-6}\) and T\_0=5 was set, creating a dual learning cycle which accelerated the convergence.\\[0.1cm]

%% file: chapters/chapter_4.tex
 The initial comparison between the models was held based on the overall
precision, recall and total time taken for the training process.
\\[0.1cm]

\begin{table}[ht]
\centering
\resizebox{\textwidth}{!}{%
\begin{tabular}{|l|c|c|c|c|}
\hline
 \textbf{} & \textbf{Model} & \textbf{Precision} & \textbf{Recall} & \textbf{Total Training Time (Hour)} \\
\hline
\multirow{2}{*}{\textbf{Classification model}} 
  & ZF- Net   & 10.93 & 16.29 & 0.65 \\
  & EfficientNetV2 & 92.52 & 91.45 & 1.6  \\
\hline
\multirow{2}{*}{\textbf{Object Detection Model}} 
  & YOLOv11 & 26.48 & 29.80 & 1.3  \\
  & F-RCNN  & 68.07 & 72.12 & 3.34 \\
\hline
\end{tabular}
}
\caption{Base Model Evaluation: Comparing results of  ZF-Net, EfficientNetV2, YOLOv11 and F-RCNN on \textbf{raw test set}}
\label{base raw}
\end{table}

\begin{table}[ht]
\centering
\resizebox{\textwidth}{!}{%
\begin{tabular}{|l|c|c|c|c|}
\hline
 \textbf{} & \textbf{Model} & \textbf{Precision} & \textbf{Recall} & \textbf{Total Training Time (Hour)} \\
\hline
\multirow{2}{*}{\textbf{Classification model}} 
  & ZF- Net   & 11.98 & 16.45 & 0.65 \\
  & EfficientNetV2 & 86.07 & 82.90 & 1.6  \\
\hline
\multirow{2}{*}{\textbf{Object Detection Model}} 
  & YOLOv11 & 25.74 & 28.58 & 1.3  \\
  & F-RCNN  & 60.96 & 64.67 & 3.34 \\
\hline
\end{tabular}
}
\caption{Base Model Evaluation: Comparing results of  ZF-Net, EfficientNetV2, YOLOv11 and F-RCNN on \textbf{augmented test set}}
\label{base aug}
\end{table}

\begin{table}[ht]
\renewcommand{\arraystretch}{1.5}
\centering
\resizebox{\textwidth}{!}{%
\begin{tabular}{|l|c|c|c|c|c|c|c|}
\hline
\textbf{Model} & \textbf{Bear mAP} & \textbf{Deer mAP} & \textbf{Fox mAP} & \textbf{Frog mAP} & \textbf{Hedgehog mAP} & \textbf{Tiger mAP} & \textbf{Turtle mAP} \\
\hline
YOLOv11 & 0.2358 & 0.4132 & 0.0222 & 0.2944 & 0.3964 & 0.0000 &  0.0000 \\
F-RCNN  & 0.5206 & 0.5301 & 0.6477 & 0.6658 & 0.5628 & 0.8996 & 0.2879 \\
\hline
\end{tabular}
}
\caption{Classwise mAP evaluation of FRCNN and YOLOv11 on \textbf{raw test set}}
\label{base map raw}
\end{table}

\begin{table}[ht]
\renewcommand{\arraystretch}{1.5}
\centering
\resizebox{\textwidth}{!}{%
\begin{tabular}{|l|c|c|c|c|c|c|c|}
\hline
\textbf{Model} & \textbf{Bear mAP} & \textbf{Deer mAP} & \textbf{Fox mAP} & \textbf{Frog mAP} & \textbf{Hedgehog mAP} & \textbf{Tiger mAP} & \textbf{Turtle mAP} \\
\hline
YOLOv11 & 0.2476 & 0.4024 & 0.0111 & 0.2877 & 0.3333 & 0.0110 & 0.0000 \\
F-RCNN  & 0.5032 & 0.4211 & 0.5213 & 0.6254 & 0.4416 & 0.8351 & 0.2576 \\
\hline
\end{tabular}
}
\caption{Classwise mAP evaluation of FRCNN and YOLOv11 on \textbf{augmented test set}}
\label{base map aug}
\end{table}

In tables \ref{base raw} and \ref{base aug}, the precision, recall and total time were noted with a batch size of 8 for Faster R-CNN, YOLOv11, EfficientNetV2, and ZFnet in both raw and augmented test sets. The classification models achieved significantly better results. However, those were not considered, as the research focused on models that were capable of object detection with proper bounding boxes. Between the two object detection models, Faster R-CNN achieved a noteworthy result with a high precision of 68.07\% and a recall of 72.12\% on the raw test set and a precision of 60.96\% and a recall of 64.67\% evaluated on the augmented test set. Faster R-CNN's two-stage architecture with RPN, ROI pooling and Resnet50 as backbone for detailed feature extraction, made it suitable for precise object detection and performed consistently well in detecting animals from trap images in both test sets respectively in comparison with YOLOv11 (table \ref{base map raw} and \ref{base map aug}). Hence, it was chosen as the base object detection model due to its reliability and accuracy in detecting animals. \\[0.1cm] 

After that, the FRCNN with Resnet50 backbone was reassessed using a batch size of 32 to ensure consistency for further comparison and it performed well in detecting animals from images. To address the limitations of the ResNet 50 architecture, the transformer architecture was explored as the backbone instead, to introduce self-attention mechanism for extracting finer details which is crucial for smaller animals. Initially, ViT was implemented as a feature extractor with FRCNN architecture, and later it was replaced by a Swin Transformer backbone combined with FPN for feature fusion. Lastly, for further enhancement FPN was replaced with Bi-FPN which is the final proposed architecture of this research. The swin transformer backbone with bidirectional feature fusion and F-RCNN as the base model, preserved local features along with global attention mechanism and accomplished an overall robust framework for object detection.   \\[0.1cm]

\section{Object Detection Task Evaluation}

In table \ref{all aug} and \ref{all raw}, comparisons among four different architectures based on both the raw and augmented test sets, in terms of precision, recall and mAP is presented. The ViT+FRCNN model performs poorly compared to others, with precision at 0.1121 for both raw and augmented test sets, while recall drops from 0.1476 (raw) to 0.1137 (augmented) and mAP declines from 0.1358 (raw) to 0.1129 (augmented). The class-wise mAP result from table \ref{map raw} shows that, the model detects only three classes (Deer, Frog, Tiger) on the raw test set, with low mAPs (0.0385, 0.1218, 0.1201), which are failing to reach the acceptable benchmark, while smaller classes remain undetected. Class-wise performance deteriorates further when evaluated on the augmented test set, showing no detection for the "Deer" class \ref{map aug}. These observations can be explained by the underlying mechanisms of ViT: ViT converts the images into fixed patches of 16x16 ratios, and treat these patches as sequences of tokens similar to natural language processing. After this, using the self attention mechanism, ViT calculates the attention weights that determine how much influence the patch should have on feature representations. This method excels in capturing the global context, however, it lacks the ability to focus on fine-grained localized features that are essential for animal detection, such as a tiger’s stripes. This critical limitation explains ViT’s underperformance in animal detection. \\[0.1cm]

\begin{table}[ht]
\renewcommand{\arraystretch}{1.5}
\centering
\small
\resizebox{0.8\textwidth}{!}{%
\begin{tabular}{|l|c|c|c|}
\hline
\textbf{Object Detection Network} & \textbf{Precision} & \textbf{Recall} & \textbf{mAP} \\
\hline
FRCNN (ResNet 50) & 0.6560 & 0.6694 & 0.6631 \\
ViT + FRCNN & 0.1121 & 0.1476 & 0.1358 \\
Swin + FPN + FRCNN & 0.8156 & 0.7915 & 0.7904 \\
Swin + Bi-FPN + FRCNN & 0.8343 & 0.8178 & 0.8161 \\
\hline
\end{tabular}
}
\caption{Performance comparison of object detection networks on \textbf{raw test set}. The Swin + Bi-FPN + Faster R-CNN configuration outperforms all others}
\label{all raw}
\end{table}

\begin{table}[ht]
\renewcommand{\arraystretch}{1.5}
\centering
\small
\resizebox{0.8\textwidth}{!}{%
\begin{tabular}{|l|c|c|c|}
\hline
\textbf{Object Detection Network} & \textbf{Precision} & \textbf{Recall} & \textbf{mAP} \\
\hline
FRCNN (ResNet 50) & 0.5983 & 0.6103 & 0.6058 \\
ViT + FRCNN & 0.1121 & 0.1137 & 0.1129 \\
Swin + FPN + FRCNN &  0.7669 & 0.7478 & 0.7476 \\
Swin + Bi-FPN + FRCNN & 0.8059 & 0.7919 & 0.7889 \\
\hline
\end{tabular}
}
\caption{Performance comparison of object detection networks on \textbf{augmented test set}. The Swin + Bi-FPN + Faster R-CNN configuration outperforms all others}
\label{all aug}
\end{table}

The Swin+FPN+FRCNN architecture exhibits better performance based on all metrics compared to both Resnet50 and ViT backbone frameworks, assessed on both test sets. The overall average precision of 0.8156, recall of 0.7915 and mAP of 0.7904 on the raw test set (table \ref{all raw}) faces a slight reduction on the augmented test set with a precision of 0.7669, recall of 0.7478 and mAP of 0.7476 (table \ref{all aug}). The architecture achieved a balanced performance relatively, in terms of the class-wise mAP results across both raw and augmented test sets compared to the previous models, especially for the Bear (0.8388 and 0.8019), Fox (0.8944 and 0.8278), Tiger (0.9000 and 0.8963) and Turtle (0.5676 and 0.4658) classes. In spite of the smaller training dataset for the Hedgehog class, this architecture shows improvement in terms of reliability. Swin transformer backbone’s hierarchical feature extraction combined with multi-level feature fusion of FPN can detect precise local details along with global context and that helps the model to improve its generalization capability which is the primary reason behind the overall improvement. \\[0.1cm] 

\begin{table}[H]
\renewcommand{\arraystretch}{1.5}
\centering
\resizebox{\textwidth}{!}{%
\begin{tabular}{|l|c|c|c|c|c|c|c|}
\hline
\textbf{Object Detection Network} & \textbf{Bear mAP} & \textbf{Deer mAP} & \textbf{Fox mAP} & \textbf{Frog mAP} & \textbf{Hedgehog mAP} & \textbf{Tiger mAP} & \textbf{Turtle mAP} \\
\hline
FRCNN (ResNet 50) & 0.5155 & 0.6868 & 0.5971 & 0.6523 & 0.5510 & 0.8502 & 0.0167 \\
ViT + FRCNN  & 0.0000 & 0.0385 & 0.0000 & 0.1218 & 0.0000 & 0.1201 & 0.0000 \\
Swin + FPN + FRCNN  & 0.8388 & 0.7917 & 0.8944 & 0.7901 & 0.5880 & 0.9000 & 0.5676 \\
Swin + Bi-FPN + FRCNN  & 0.8351 & 0.8179 & 0.9176 & 0.7402 & 0.6475 & 0.8469 & 0.4935 \\
\hline
\end{tabular}
}
\caption{Classwise mAP evaluation of object detection networks on \textbf{raw test set}}
\label{map raw}
\end{table}

\begin{table}[H]
\renewcommand{\arraystretch}{1.5}
\centering
\resizebox{\textwidth}{!}{%
\begin{tabular}{|l|c|c|c|c|c|c|c|}
\hline
\textbf{Object Detection Network} & \textbf{Bear mAP} & \textbf{Deer mAP} & \textbf{Fox mAP} & \textbf{Frog mAP} & \textbf{Hedgehog mAP} & \textbf{Tiger mAP} & \textbf{Turtle mAP} \\
\hline
FRCNN (ResNet 50) & 0.4223 & 0.6088 & 0.5156 & 0.6505 & 0.4878 & 0.8041 & 0.0000 \\
ViT + FRCNN  & 0.0000 & 0.0000 & 0.0000 & 0.0783 & 0.0000 & 0.1156 & 0.0000 \\
Swin + FPN + FRCNN  & 0.8019 & 0.7083 & 0.8278 & 0.7500 & 0.5517 & 0.8963 & 0.4658 \\
Swin + Bi-FPN + FRCNN  & 0.8213 & 0.7037 & 0.8385 & 0.7513 & 0.6163 & 0.8401 & 0.4459 \\
\hline
\end{tabular}
}
\caption{Classwise mAP evaluation of object detection networks on \textbf{augmented test set} }
\label{map aug}
\end{table}

The proposed architecture exhibits the most optimal result considering both the testset with an average precision of 0.8343, recall of 0.8178 and 0.8161 mAP on raw test set (table \ref{all raw}) and precision of 0.8059, recall of 0.7919 and mAP of 0.7889 for augmented test set (table \ref{all aug}). The class-wise mAP for raw test set (table \ref{map raw}) demonstrates consistent improvement across classes, with a notable increase for the Fox class (0.9176). The Turtle class achieves low performance with mAP of 0.4935 and 0.4459 respectively for raw and augmented test sets. The bidirectional weighted feature fusion facilitates detection of complex patterns and optimal detection capability combined with the transformer backbone which handles the shortcomings of the previously discussed architectures. \\[0.1cm]

\begin{figure}[H]
\centering
\includegraphics[width=\textwidth]{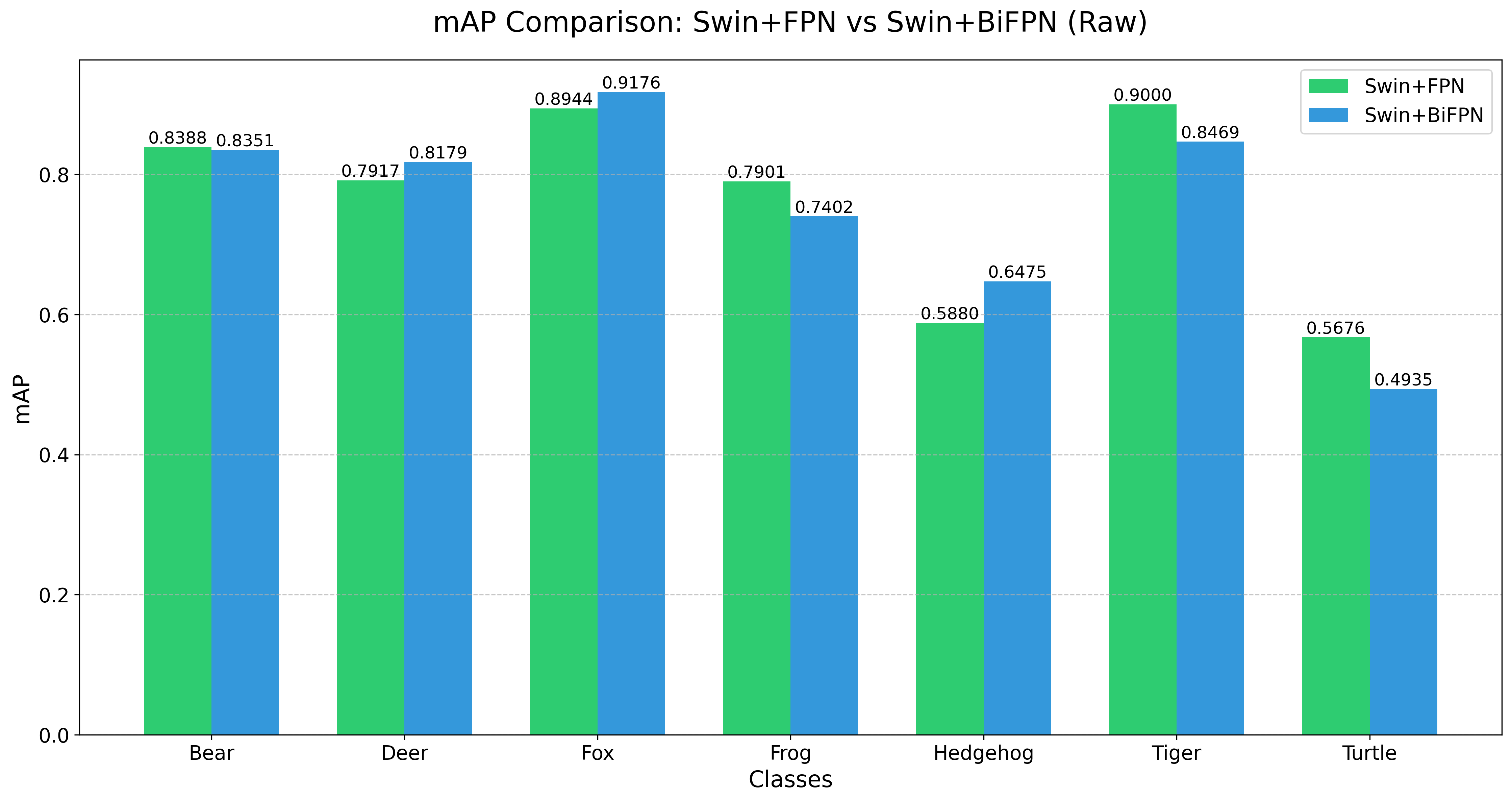}
\caption{Bar chart of classwise mAP comparison of swin+FPN and swin+BiFPN backbone on \textbf{raw test set}} \label{raw_map_fig}
\end{figure} 

\begin{figure}[H]
\centering
\includegraphics[width=\textwidth]{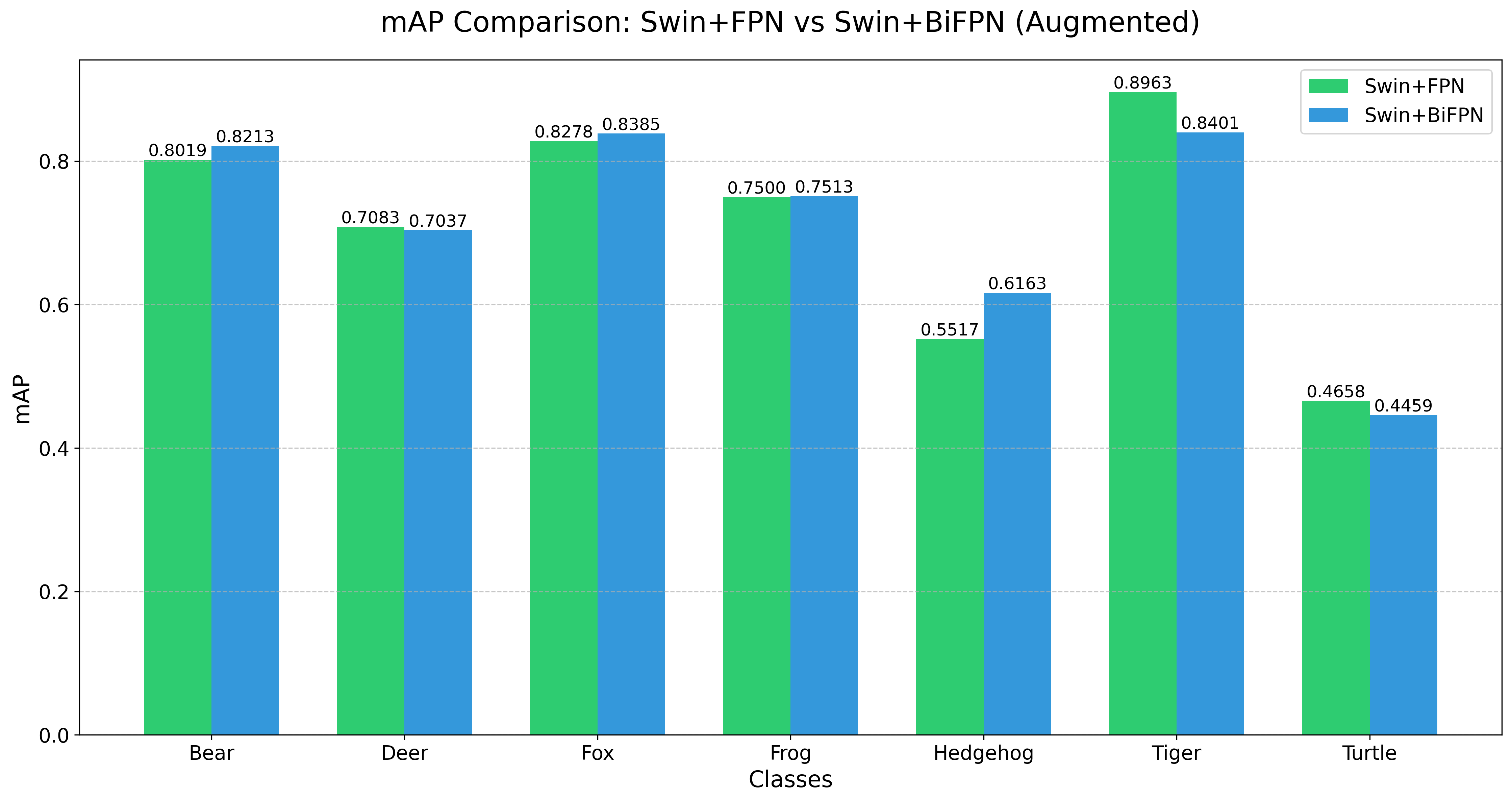}
\caption{Bar chart of classwise mAP comparison of swin+FPN and swin+BiFPN backbone on \textbf{augmented test set}} \label{aug_map_fig}
\end{figure} 

However, there are some key observations regarding the performance of the turtle and tiger classes. While the Swin Transformer combined with the FPN backbone achieves relatively higher mAP scores for these classes: 0.5676 (raw) and 0.4658 (augmented) for turtle, and  0.9000 (raw) and 0.8963 (augmented) for tiger, the Swin+BiFPN architecture demonstrates a decline in performance. The turtle class records a lower mAP of 0.4935 on the raw test set and 0.4459 on the augmented test set, whereas the tiger class records a lower mAP of 0.8469 on the raw test set and 0.8401 on the augmented test set (Figure \ref{raw_map_fig} and \ref{aug_map_fig}). This decline could be attributed to a key factor that suggests that the Swin+BiFPN model needs more training epochs to properly learn the distinguishing features extracted by the backbone. Since we had restricted access to resources and there was a time constraint, the model was trained on a lower number of epochs of 10, potentially preventing the network from fully converging and learning more discriminative features. While this would not affect the  Swin+FPN backbone, this could have an effect on the Swin+BiFPN backbone due to more fine-grained feature extraction. \\[0.1cm]

Furthermore, the overall performance of the Swin+BiFPN backbone surpasses the Swin+FPN architecture. In the augmented test set, Swin+BiFPN outperforms Swin+FPN in four out of seven classes (Figure \ref{aug_map_fig}). This leads to two key observations: First, the number of training epochs required for the model to learn richer feature representations may have influenced the performance of the Swin+BiFPN backbone and second, the similarity between augmented test images and the training data may contribute positively to the model’s effectiveness in the augmented test set.
\\[0.2cm]

\begin{figure}[H]
\centering
\includegraphics[width=0.8\textwidth]{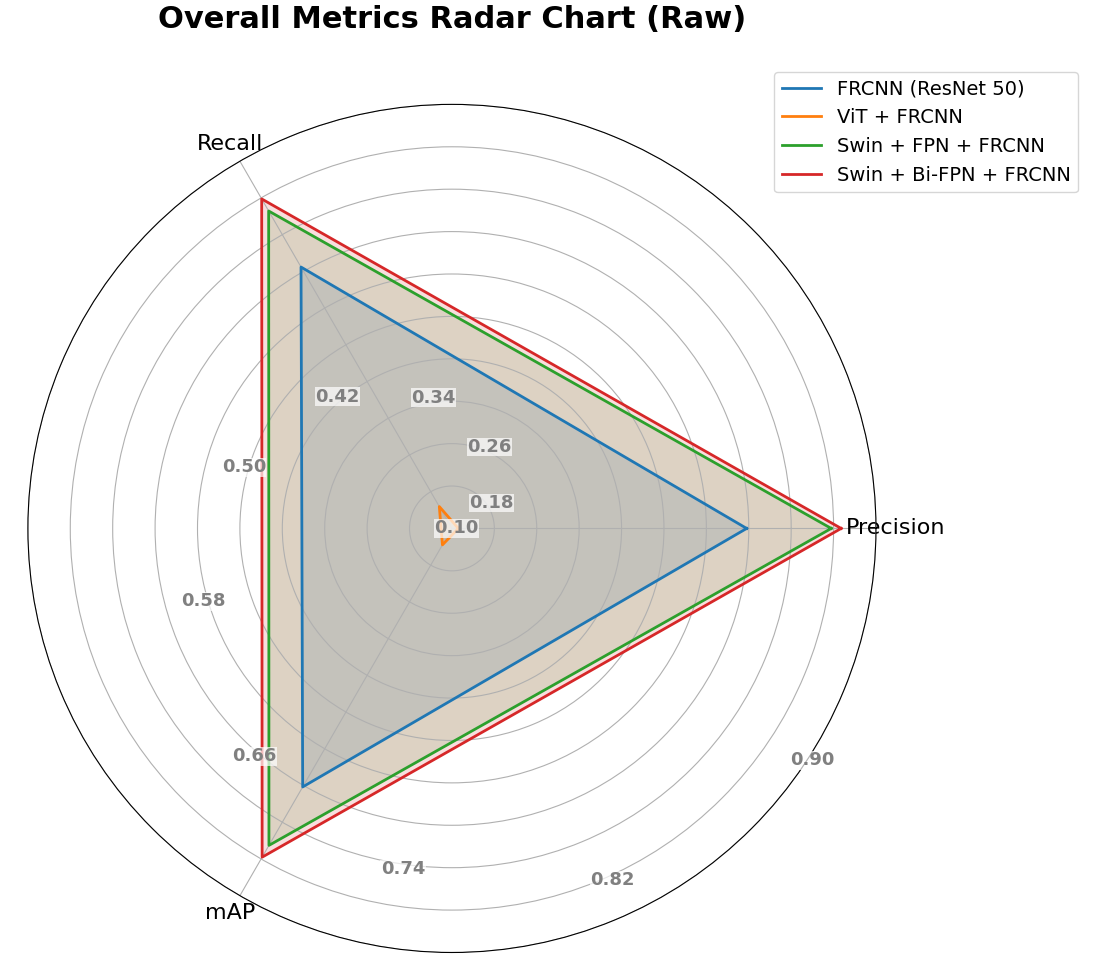}
\caption{Performance comparison visualization of object detection networks on \textbf{raw test set} using radar chart} \label{raw_radar_fig}
\end{figure} 

\begin{figure}[H]
\centering
\includegraphics[width=0.8\textwidth]{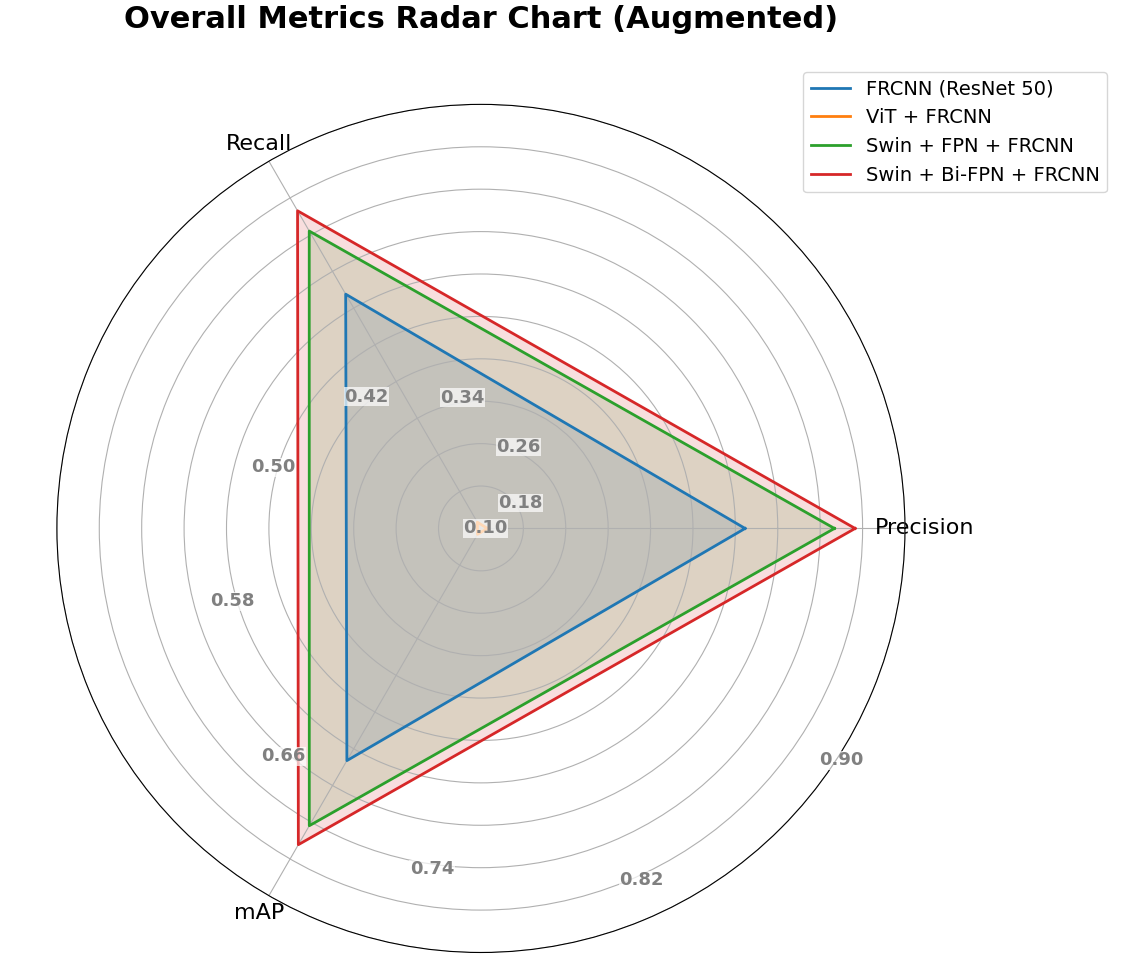}
\caption{Performance comparison visualization of object detection networks on \textbf{augmented test set} using radar chart} \label{aug_radar_fig}
\end{figure}

The results presented in this section demonstrate that the proposed object detection framework significantly outperforms traditional object detection models across both the raw and augmented test datasets. Conventional CNN-based object detectors show limited effectiveness in complex and cluttered environments, as found in trap images. In contrast, the integration of a self-attention-based backbone with Bi-FPN for enhanced feature extraction leads to substantial improvements in detection accuracy under challenging environmental conditions. This conclusion is further illustrated in the figures \ref{raw_radar_fig} and \ref{aug_radar_fig} where the radar charts show that Swin + Bi-FPN + FRCNN (red) achieves the highest values across all three metrics of precision, recall and mAP, closely followed by Swin + FPN + FRCNN (green) while CNN based model ResNet50 demonstrates lower performance (blue) in both raw and augmented test sets. \\[0.05cm]
Notably, despite being trained on a relatively small augmented dataset, the proposed model maintains competitive performance when evaluated on entirely different test images. Since the test images were taken from different datasets, they varied significantly from the images the model was trained on. This observation strongly demonstrates the model’s strong generalization capability which is essential for practical real-world scenarios. The results suggest that with access to larger and more diverse trap image datasets, the model's generalization power could be further enhanced, making it an even more robust solution for wildlife conservation.\\[0.05cm]
Furthermore, the architecture shows promising results in detecting small and partially occluded animals. This capability is particularly critical in wildlife conservation efforts, where traditional models often fall short. Overall, the experimental findings validate the effectiveness of the proposed object detection architecture and demonstrate how attention-based backbones are not only efficient but also achieves higher performance rates.\\[0.1cm]

\section{Visual Semantics Extraction Task Evaluation}

After successful animal detection using our proposed Swin-BiFPN-FRCNN network, the next stage of the pipeline involves visual semantic extraction. As discussed earlier, this enables the architecture to introduce zero-shot detection capabilities, as well as providing important and emergent cues of the animal’s behavior, further advancing the conservation effort. \\[0.1cm]

\textbf{Zero-Shot Detection:} Zero-shot detection refers to the task where a model can identify and localize objects of a category without any prior training examples. This contrasts with the traditional object detection, where models are trained on labeled data for each class they are expected to detect. \\[0.1cm]

To evaluate the effectiveness of the generated description, five Multimodal Large Language Models (MLLMs) were assessed. These include LlaVa v1.5 7B, LlaVa v1.5 13B, LlaVa v1.6 Mistral, KOSMOS 2 and IDEFICS instruct 9B. Each of the five MLLMs evaluated in this study is publicly available as an open-source model on the Hugging Face Hub. Although BLIP-2 was initially included in the evaluation, it consistently failed to generate meaningful or contextually relevant descriptions. As a result, the model was excluded from further analysis. \\[0.1cm]

To quantitatively assess the semantic relevance and accuracy of the generated descriptions, we have followed a dual-approach: a) Textual similarity analysis using BERTScore and SBERT cosine similarity against the ground truth caption. b) LLM as a Judge: where two strong language models, GPT-4.1 and GROK 3.0, acted as a ‘judge’ to compare the different outputs and generate score based on relevance, accuracy, depth, and fluency. This dual evaluation strategy enables assessment of the captions relative to ground-truth references as well as through direct comparison between model outputs. \\[0.1cm]

\subsection{Textual Similarity Results}

\begin{table}[h!]
\centering
\renewcommand{\arraystretch}{1.3}
\resizebox{\textwidth}{!}{%
\begin{tabular}{|l|c|c|c|c|}

\hline
\multicolumn{1}{|c|}{\textbf{MLLM Model}} & 
\multicolumn{3}{c|}{\textbf{BERTScore}} & 
\textbf{SBERT Cosine Similarity} \\
\cline{2-5}
& \textbf{Precision} & \textbf{Recall} & \textbf{F1 Score} & \textbf{Mean Score} \\
\hline
LLaVa v1.5 7B & 0.9018 & 0.8930 & 0.8972 & 0.5788 \\
LLaVa v1.5 13B & 0.8952 & 0.9007 & 0.8978 & 0.6501 \\
LLaVa v1.6 Mistral & 0.8741 & 0.8971 & 0.8854 & 0.6283 \\
KOSMOS 2 & 0.8722 & 0.8841 & 0.8780 & 0.5513 \\
IDEFICS 9B & 0.8840 & 0.8547 & 0.8690 & 0.5718 \\
\hline
\end{tabular}
}
\caption[short title]{Evaluation of MLLM Models using BERTScore and SBERT Cosine Similarity}
\label{tab5}
\end{table}

Table \ref{tab5} summarizes the performance evaluation of the five MLLMs using BERTScore (Precision, Recall, F1 Score) and SBERT Cosine Similarity against referenced captions. The LlaVa v1.5 13B achieved the highest SBERT similarity (0.6501), indicating good semantic alignment with the ground-truth. Furthermore, the high precision (0.8952), recall (0.9007), and F1 score (0.8978) indicate the model’s ability to generate accurate and comprehensive descriptions, including key information. One concern regarding this model was its relatively higher parameter count compared to the others. However, timing tests revealed that LLaVA v1.5 (13B) required only 1.23 minutes to generate descriptions over an image, whereas LLaVA v1.6 Mistral, despite having fewer parameters (7B), took 2 minutes to complete.\\[0.1cm]

\subsection{LLM-as-a-Judge}

To ensure comprehensive and contrastive evaluation across all five MLLMs, we also employed the LLM-as-a-judge evaluation approach. The ‘judge’ models’ extensive parameters allow it to process and compare all five model outputs concurrently. This approach allows both qualitative and quantitative evaluation of the generated descriptions. \\[0.1cm]

\begin{table}[h!]

\centering
\renewcommand{\arraystretch}{1.3}
\resizebox{\textwidth}{!}{%
\begin{tabular}{|l|c|c|c|c|c|}
\hline
\textbf{MLLM model} & \textbf{Avg Relevance} & \textbf{Avg Accuracy} & \textbf{Avg Depth} & \textbf{Avg Fluency} & \textbf{Total} \\
\hline
LLaVa v1.5 7B & 0.6043 & 0.6429 & 0.6184 & 0.9836 &2.8491 \\
LLaVa v1.5 13B &0.65 & 0.8114 & 0.7484 & 0.9707 & 3.1806 \\
LLaVa v1.6 Mistral & 0.6286 & 0.5686 & 0.9489 & 0.9986 & 3.1446 \\
KOSMOS 2 & 0.46 & 0.81 & 0.7953 & 0.4993 & 2.5646 \\
IDEFICS 9B & 0.5643 & 0.0071 & 0.3479 & 0.3986 & 1.3179 \\
\hline
\end{tabular}
}
\caption{LLM-as-a-Judge evaluation using GPT 4.1}
\label{gpt}
\end{table}

\begin{table}[h!]

\centering
\renewcommand{\arraystretch}{1.3}
\resizebox{\textwidth}{!}{%
\begin{tabular}{|l|c|c|c|c|c|}
\hline
\textbf{MLLM model} & \textbf{Avg Relevance} & \textbf{Avg Accuracy} & \textbf{Avg Depth} & \textbf{Avg Fluency} & \textbf{Total} \\
\hline
LLaVa v1.5 7B & 0.85 & 0.80 & 0.75 & 0.82 & 3.22 \\
LLaVa v1.5 13B & 0.92 & 0.88 & 0.85 & 0.90 & 3.55 \\
LLaVa v1.6 Mistral & 0.88 & 0.84 & 0.80 & 0.87 & 3.39 \\
KOSMOS 2 & 0.90 & 0.82 & 0.70 & 0.80 & 3.22 \\
IDEFICS 9B & 0.87 & 0.83 & 0.78 & 0.89 & 3.37 \\
\hline
\end{tabular}
}
\caption{LLM-as-a-Judge evaluation using GROK 3.0}
\label{grok}
\end{table}

The models selected for this approach are GPT-4.1 and GROK 3.0. Tables \ref{gpt} and \ref{grok} present the evaluation results from each judge, respectively. GPT 4.1 is an advanced state-of-the-art language model with strong contextual comprehension, and GROK 3.0 features advanced reasoning capabilities, making them both well-suited for the evaluation. The judges were prompted \hyperref[sec:prompt]{(Go to Prompt)} to assess the outputs from all five models scoring each on a scale from 0 to 1 across four metrics: relevance, accuracy, depth, and fluency and then the scores were combined to provide an overall total. Based on the evaluations from both judges, LLaVA v1.5 (13B) achieved the highest scores, receiving a total of 3.1806 from GPT-4.1 and 3.55 from GROK 3.0. These results further support the BERTScore and SBERT findings. To mitigate potential biases from the LLM judges, the order of the output files was randomized during evaluation. \\[0.1cm]

\begin{figure}[H]
  \centering
  \begin{minipage}[t]{0.45\textwidth}
    \centering
    \includegraphics[width=\linewidth]{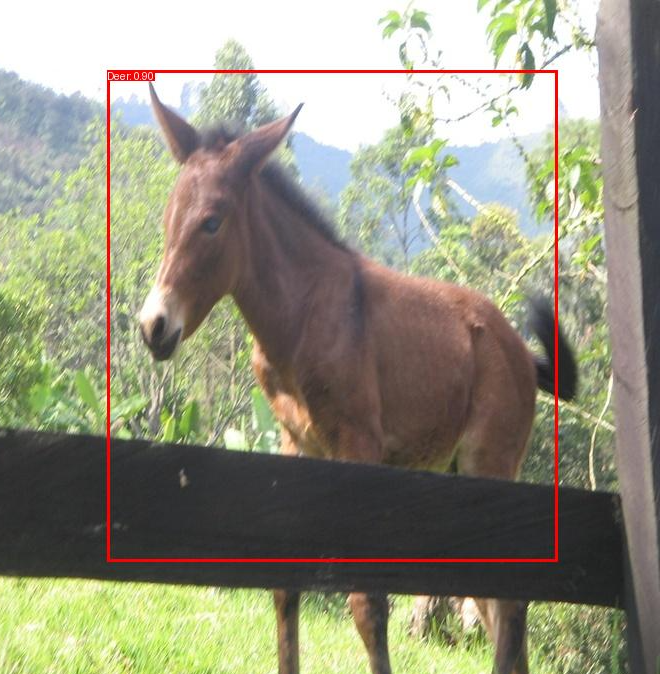}
    \label{horse}
  \end{minipage}%
  \hfill
  \begin{minipage}[t]{0.45\textwidth}
    \centering
    \includegraphics[width=\linewidth]{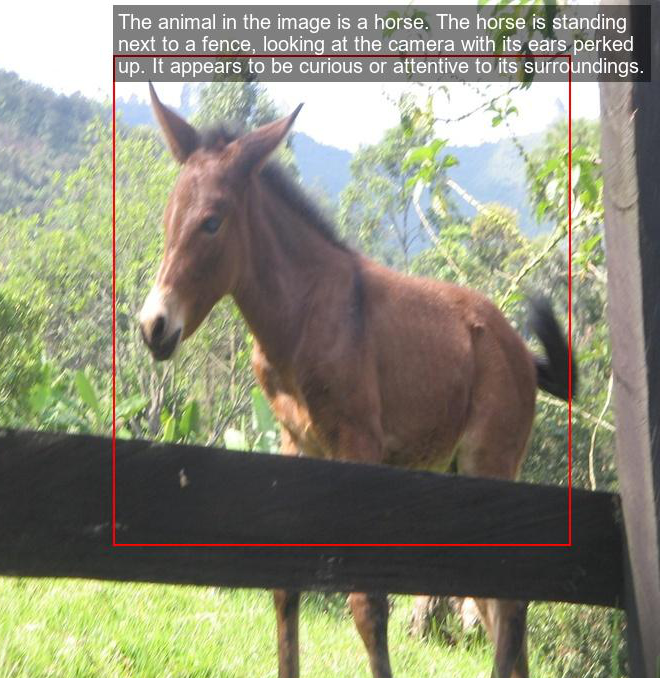}
    \label{horse detected}
  \end{minipage}

  \vspace{0.5cm}
  \caption{Zero-shot detection: While the object detection network misclassified a horse as a deer due to lack of training data, the MLLM correctly identified it, highlighting the complementary strength of vision-language models in refining detection outcomes as a fallback option}
  \label{fig:overall_horse_comparison}
\end{figure}

Figure \ref{fig:overall_horse_comparison} illustrates a scenario where the detection module fails to correctly classify an image belonging to a category it was not trained on. In contrast, the MLLM successfully identifies the object, demonstrating the framework's zero-shot capability and highlighting the enhanced robustness achieved through the integration of the MLLM. This integration serves as a complementary fallback mechanism to the detection network, enhancing overall system reliability in unfamiliar scenarios.\\[0.1cm]

To summarize, the presented architecture in this research demonstrates a detailed framework for wildlife conservation and management efforts with enhanced animal detection capabilities along with meaningful textual descriptions. The proposed object detection model (Swin+Bi-FPN+F-RCNN) obtained a remarkable result with a high precision of 0.8343 (raw) and 0.8059 (augmented) and recall of 0.8178 (raw) and 0.7919 (augmented), which indicates strong generalization capability and promising detection capabilities in low contrast environments and it is indispensable for conservation strategies such as population count. Also the model managed to detect small animals with a relatively high mAP in complex backgrounds which is essential for wildlife scenarios. The implementation of MLLM model, particularly LLaVA-1.5-13b-hf, achieves superior performance (F1 score: 0.8978), generates meaningful and relevant descriptions of the species, leverages zero-shot detection capabilities and which is vital for monitoring unknown endangered wildlife species. Therefore, this comprehensive pipeline fuses object detection with adequate language modeling ability that introduces a functional, efficient and reliable framework for wildlife conservation, leading to improved monitoring systems for preservation. \\[0.1cm]

\section{Limitations and Future Work}
While this study demonstrates how the proposed framework can contribute to successful wildlife conservation, some limitations must be acknowledged that present opportunities for further enhancement. Firstly, due to time retrains and limited access to resources, the models were trained for a relatively lower number of epochs. This may explain the Swin+BiFPN model's reduced ability to effectively learn and generalize fine-grained features. Secondly, the dataset was augmented to mimic the trap image-like challenges due to the unavailability of trap images. Hence, the dataset was limited and smaller, which may have affected the model's performance. Furthermore, although the framework shows some capabilities of zero-shot detection, using the MLLM model, the detection framework still relies on a trained dataset. \\[0.1cm]

The future work for this study includes several key directions aiming at enhancing the effectiveness of the model. First, integrating object segmentation alongside detection can further enhance the model's feature learning capability. Second, introducing zero-shot capability in the overall framework by leveraging LLM models within the object detection module would allow the system to be more robust by enabling the system to detect and describe previously unseen object categories without requiring additional training data. Lastly, optimizing inference speed to further support the conservation effort at a low computational cost.
\\[0.1cm]

%% file: chapters/chapter_5.tex
This thesis presents a comprehensive and robust framework for wildlife animal detection in camera trap images, focusing on reducing the limitations in the current approaches. By integrating Swin transformer along with Bi-Directional Feature Pyramid Network in the Faster RCNN detection network, along with a visual semantic extraction module utilizing LLaVA v1.5 (13B), the system achieves high detection accuracy, particularly in low-contrast environments, while demonstrating effective generalization abilities. Additionally, the incorporation of the Multimodal Large Language Model enables zero-shot detection as well as enriches outputs with meaningful behavioral insights, contributing to conservation efforts. Evaluation through both traditional NLP metrics and LLM-based judges confirms the robustness of the framework. Overall, this study lays the foundation of an LLM-based transformer detection framework approach, reducing manual workload while monitoring wildlife population and conservation efforts.

%% file: chapters/chapter_7.tex
You are an expert in animal behavior and explanation clarity.\\[0.1cm]

Five models were asked to describe what the animal is doing and why. \\[0.1cm]

Please read the responses and rate each from 0 to 1 (continuous) in the following categories:\\[0.1cm]
Relevance (is it about the image?)\\[0.05cm]
Biological accuracy (plausible behavior?)\\[0.05cm]
Explanation depth (goes beyond surface?)\\[0.05cm]
Fluency (clear and coherent?)\\[0.1cm]

Return the results in this JSON format:\\[0.1cm]
\{
  "LlaVA  1.5 13b hf": \{ "relevance": , "accuracy": , "depth": , "fluency": , "total":  \},\\[0.1cm]
  "LlaVA 1.5 7b hf": \{ "relevance": , "accuracy": , "depth": , "fluency": , "total":  \},\\[0.1cm]
  "LlaVA 1.6 7b Mistral": \{ "relevance": , "accuracy": , "depth": , "fluency": , "total":  \}, \\[0.1cm]
  "Kosmos 2": \{ "relevance": , "accuracy": , "depth": , "fluency": , "total":  \},\\[0.1cm]
  "Idefics 9b Instruct": \{ "relevance": , "accuracy": , "depth": , "fluency": , "total":  \},\\[0.1cm]
 "winner": "..." 
\}\\[0.1cm]

Check all 700 responses for each model and give me average of each metric. 
Please evaluate.